\documentclass{article}

\usepackage{microtype}
\usepackage{graphicx}
\usepackage{booktabs} % for professional tables

\usepackage{hyperref}
\usepackage{multirow}            % tabular cells spanning multiple rows
\usepackage{amsfonts}            % blackboard math symbols
\usepackage{duckuments}          % sample images

\usepackage{subfig}
\usepackage{tablefootnote}
\usepackage{xcolor,colortbl}
\usepackage{mathtools}
\usepackage{amsmath}
\usepackage{amsbsy}
\usepackage{amssymb}
\usepackage{pdflscape}

\usepackage{times}
\usepackage{soul}
\usepackage{url}
\usepackage{caption}
\usepackage{graphicx}
\usepackage{amsmath}
\usepackage{amsthm}
\usepackage{booktabs}

\usepackage[switch]{lineno}
\newcommand\sitao[1]{\textcolor{red}{[Sitao: #1]}}
\input{env}
\usepackage[accepted]{icml2024}

\usepackage{amsmath}
\usepackage{amssymb}
\usepackage{mathtools}
\usepackage{amsthm}

\usepackage[capitalize,noabbrev]{cleveref}

\theoremstyle{plain}

\theoremstyle{definition}
\newtheorem{definition}[theorem]{Definition}

\theoremstyle{remark}

\usepackage[textsize=tiny]{todonotes}

\icmltitlerunning{Representation Learning on Heterophilic Graph with Directional Neighborhood Attention }

\begin{document}

\twocolumn[
\icmltitle{Representation Learning on Heterophilic Graph with Directional Neighborhood Attention }



\icmlsetsymbol{corresponding}{$\dagger$}

\begin{icmlauthorlist}
\icmlauthor{Qincheng Lu}{mcgill}
\icmlauthor{Jiaqi Zhu}{mcgill}
\icmlauthor{Sitao Luan}{corresponding,mcgill,mila}
\icmlauthor{Xiao-Wen Chang}{corresponding,mcgill}
\end{icmlauthorlist}

\icmlaffiliation{mcgill}{McGill University}
\icmlaffiliation{mila}{Mila - Quebec Artificial Intelligence Institute}

\icmlcorrespondingauthor{Sitao Luan}{sitao.luan@mail.mcgill.ca}
\icmlcorrespondingauthor{Xiao-Wen Chang}{chang@cs.mcgill.ca}

\icmlkeywords{Machine Learning, ICML}

\vskip 0.3in
]



\printAffiliationsAndNotice{\icmlCorrespondingAuthor} % otherwise use the standard text.

\begin{abstract}
Graph Attention Network (GAT) is one of the most popular Graph Neural Network (GNN) architecture, which employs the attention mechanism to learn edge weights and has demonstrated promising performance in various applications. However, since it only incorporates information from immediate neighborhood, it lacks the ability to capture long-range and global graph information, leading to unsatisfactory performance on some datasets, particularly on heterophilic graphs. To address this limitation, we propose the Directional Graph Attention Network (DGAT) in this paper. DGAT is able to combine the feature-based attention with the global directional information extracted from the graph topology. To this end, a new class of Laplacian matrices is proposed which can provably reduce the diffusion distance between nodes. Based on the new Laplacian, topology-guided neighbour pruning and edge adding mechanisms are proposed to remove the noisy and capture the helpful long-range neighborhood information. Besides, a global directional attention is designed to enable a topological-aware information propagation. The superiority of the proposed DGAT over the baseline GAT has also been verified through experiments on real-world benchmarks and synthetic data sets. It also outperforms the state-of-the-art (SOTA) models on 6 out of 7 real-world benchmark datasets.
\end{abstract}

\section{Introduction}

Combining graph signal processing and Convolutional Neural Networks (CNNs) \cite{lecun1998gradient}, Graph Neural Networks (GNNs) has achieved remarkable results on machine learning tasks with non-Euclidean data \cite{scarselli2008graph,kipf2016semi,
hamilton2017inductive,velivckovic2017graph,xu2018powerful,luan2019break,beaini2021directional,zhao2021consciousness,hua2023mudiff}. Unlike CNNs where neighbours are weighted differently, the convolutional kernel in the vanilla GNNs \cite{kipf2016semi} and many other popular variants \cite{hamilton2017inductive, xu2018powerful} assign the same weight in a neighbourhood. The Graph Attention Network (GAT) \cite{velivckovic2017graph} overcomes this limitation by allowing nodes to connect to its local neighbours with learnable weights.
%The self-attention mechanism \cite{vaswani2017attention} employed in the graph domain by GAT contributes to its success. 

The GAT model generates the node representation solely based on the representations of its direct neighbours and this works well on the homophilic graphs \cite{luan2022we}, where the node and its neighbours are likely to have the same label. However, GAT suffers from significant performance loss in node classification tasks involving heterophilic graphs \cite{zhu2020beyond,luan2022complete}, where nodes from different classes tend to be connected.  

Non-local neighborhood information is found to be helpful to deal with heterophily problem for GNNs \cite{abu2019mixhop,liu2020non,zhu2020beyond,he2021bernnet,zheng2022graph}, but it has not been leveraged in GAT model and we will address this issue in this paper. The main contribution of our paper is: 
(1) We propose a new class of parameterized normalized graph Laplacian matrices which have better control over directional aggregation and can contain the widely used normalized Laplacians as special cases. %and whose eigenvalues lie in $[0,2]$.
%and provides a systematic way to control its eigenfunctions;
%whose first non-trivial eigenvalue adapts the message passing 
%from which we can reconstruct a new adjacency matrix \red{Should it be a matrix with parameters?} to capture long-range neighborhood information to alleviate heterophily problem. 
Through the parameters in new Laplacian, we can adjust the diffusion distances and spectral distances between nodes by altering the spectral property of the graph;
% \red{(1) and (2) are about the same thing. They should be merged.
% How about the contribution of Theorem 3.3?
% It's purpose is to have 2 parameters which can be adjusted.
% YOu didn't mention Them 3.3.}
%\blue{I think only two points is needed}
%\sitao{Through the new Laplacian, we can theoretically reduce the diffusion distance between ..., thus, it can provably capture the global information ... }. '=
(2) We establish a theorem that enables the usage of the spectral distance as a surrogate function of diffusion distance, which significantly reduce the computational cost for comparing the relative distance between nodes. It substantially extends the scope and overcomes some shortcomings of \cite{beaini2021directional}.
%\red{This is not enough. Look at the significance in the text.
% You need to say how good the results are.
% I mean Theorem 3.3. There are comments around that theorem.
% Summarize them and put it here.
% Why don't you want to say more? Look at footnote 3.}
(3) Based on the new graph Laplacian, 
we propose the Directional Graph Attention Network (DGAT).
The contributions of DGAT are two folds: 
i. a topology-guided graph rewiring strategy with theoretical justifications;
ii. a global directional attention which enables a topological-aware information propagation.
%\sitao{ which prune or rewire according to ... so that ...}  
%which performs graph rewiring according to the node relative distance determined by the proposed graph Laplacian,
%and enables a global topological-aware information propagation through an enhanced graph attention mechanism. 
%\sitao{The results...}
The empirical results demonstrate the effectiveness and superiority of DGAT compared with SOTA GNNs for node classification tasks on graph across various homophily levels.
The proposed strategy is characterized by flexibility and seamless integration with the original GAT architecture and other types of graph attention mechanism.

This paper will be organized as follows:  notation and background knowledge are introduced in Section \ref{sec:preliminaries}; in Section \ref{sec:DAM}, we propose the new class of Laplacian, show its properties, prove the theoretical results on spectral and diffusion distances and put forward the DGAT architecture; in Section \ref{sec:related_works}, we discuss related works; in Section \ref{sec:experiments}, we show the experimental results and comparisons on synthetic and real-world graphs.

%\sitao{This paper will be organized as follows: in Section ..., we ...; in Section, we ...}
% during the message passing, 
% which aims to alleviate the aforementioned issue 
% that the original GAT model has by exploiting both the local feature-based information and the global topology-based directional flow.
% To be more specific, the process of local feature-based message aggregation is achieved by employing the standard attention mechanism in GAT. 
% As a consequence, our primary focus remains on devising the global topology-based aggregation strategy. 

\section{Preliminaries}
\label{sec:preliminaries}

In this section, we introduce notation and background knowledge.
We use {\bf bold} font for vectors and matrices.
For a matrix $\B=(b_{ij})$, $|\B|=(|b_{ij}|)$
and its $i^{th}$ row is denoted as $\B_{i,:}$ or $\mathbf{b}_i^\top$.
We use $\B || \C$ and $[\B, \C]$ to denote column and row concatenation of matrices $\B$ and $\C$, respectively.
We use $\mathcal{G}=(\mathcal{V}, \mathcal{E})$ to denote an undirected connected graph with the vertex set $\mathcal{V}$ (with $|\mathcal{V}| = N$) and the edge set $\mathcal{E}$.
We have a node feature matrix $\mathbf{X} \in \Rbb^{N \times d_{0}}$ whose $i^{th}$ row is the transpose of the feature vector $\mathbf{x}_i \in \Rbb^{d_{0}}$ of node $v_i$.
The learned node representation matrix 
at the $l^{th}$ layer of GNNs is denoted 
by $\mathbf{H}^{(l)} \in \Rbb^{N \times d_{l}}$.
% and use $\mathbf{H} \in \Rbb^{N \times d_{k}}$ to denote the node hidden representation learned by the $k^{th}$ layer of a GNN.
For a node $v_i \in \mathcal{V}$, 
$\mathcal{N}(v_i) \subseteq \mathcal{V}$ denotes the set of neighbouring nodes of $v_i$.
The adjacency matrix of $\mathcal{G}$ is denoted by $\mathbf{A}=(a_{ij}) \in \Rbb^{N \times N}$ with $a_{ij} = 1$ if $e_{ij} \in \mathcal{E}$ and $a_{ij} = 0$ otherwise.
The degree matrix of $\mathcal{G}$ is 
$\mathbf{D}=\diag(d_{ii})\in \Rbb^{N \times N}$
with $d_{ii} = \sum_{j} a_{ij}$.
Three Laplacian matrices, namely the combinatorial Laplacian, random-walk normalized Laplacian and symmetric normalized Laplacian are respectively defined as:
\be 
\label{def:lap}
\mathbf{L} = \mathbf{D} - \mathbf{A}, \ \ \ \mathbf{L}_{\mathrm{rw}} = \mathbf{D}^{-1} \mathbf{L},  \ \ \ \mathbf{L}_\mathrm{sym} = \mathbf{D}^{-\frac{1}{2}} \mathbf{L} \mathbf{D}^{-\frac{1}{2}}.
\ee
We use $\boldsymbol{\phi}^{(k)}$ to denote the $k^{th}$ eigenvector of a Laplacian corresponding to the $k^{th}$ smallest positive eigenvalue $\lambda^{(k)}$.

\subsection{Graph Neural Networks}
Most modern GNNs adopt the message-passing framework \cite{hamilton2020graph}, 
in which the representation $\mathbf{h}_u$ of node $u$ is generated by iterative aggregation of its neighbourhood representations and its own representation generated from the previous layer. The $l^{th}$ layer of a GNN can be represented by the two following operations \cite{hamilton2020graph}:
\be
\begin{aligned}
\label{def:message}
&\m^{(l)}_u = \agg^{(l)}(\{\mathbf{h}^{(l-1)}_v: \forall v \in\mathcal{N}(u)\}), \\ 
&\mathbf{h}^{(l)}_u = \update^{(l)}(\mathbf{h}^{(l-1)}_u, \m^{(l)}_u),  
\end{aligned}
\ee
where $\m^{(l)}_u$ is the aggregated message by applying the $\agg$ operator to node $u$'s direct neighbours;
and $\mathbf{h}^{(l)}_u$ is the representation for node $u$ in the $l^{th}$ layer generated by the $\update$ operator. 

\subsection{Graph Attention Mechanism}
The graph attention is defined as a function $g: u \times \mathcal{N}(u) \xrightarrow{} [0, 1]$ that maps the node $u$ and any node in $\mathcal{N}(u)$ to a relevance score \cite{lee2019attention}. 
For example, $r_{ij}^{(l)}$, the attention score that node $v_i$ gives to node $v_j$ at the $l^{th}$ layer can be defined as
\be
\label{eq:leaky_attention}
r_{ij}^{(l)} = \text{LeakyReLU}\big((\mathbf{a}^{(l)})^{\top} [\mathbf{W}^{(l)} \mathbf{h}_i^{(l-1)} || \mathbf{W}^{(l)} \mathbf{h}_j^{(l-1)}]\big).
\ee
where $\mathbf{a}^{(l)} \in \Rbb^{2 d_{l}}$ and $\mathbf{W}^{(l)} \in \Rbb^{d_{l} \times d_{l-1}}$ are learnable parameters.
Denote $\mathbf{R}^{(l)}= (r_{ij}^{(l)})\in \Rbb^{N\times N}$.
Then GAT uses a mask matrix $\mathbf{M}=(m_{ij})\in \Rbb^{N\times N}$ to preserve the structural information and normalize the attention score:
\be 
\begin{aligned}
\label{eq:attentionscore}
& \alpha_{ij}^{(l)} = \softmax_{j}(\widetilde{\mathbf{R}}^{(l)}_{i, :}), \ \ \ \widetilde{\mathbf{R}}^{(l)} =  \mathbf{R}^{(l)} + \mathbf{M},  \\
& m_{ij} = \left\{\begin{matrix}
 0, & e_{ij} \in \mathcal{E} \\
-\infty, & \text{otherwise}
\end{matrix}\right..
\end{aligned}
\ee
The representation $\mathbf{h}^{(l)}_i$ for node $v_i$ at the $l^{th}$ layer is calculated as follows:
% \be 
% \begin{aligned}
% \label{eq:attentionrep}
% \mathbf{h}_i^{(l)} = \sigma \left( \sum_{j: v_j \in \{ \mathcal{N}(v_i), v_i\})}  \alpha_{ij}^{(l)} \mathbf{W}^{(l)}  \mathbf{h}_j^{(l-1)} \right),
% \end{aligned}
% \ee
\be 
\begin{aligned}
\label{eq:attentionmp}
&\m^{(l)}_i = \sum_{j: v_j \in \{ \mathcal{N}(v_i)\}}  \alpha_{ij}^{(l)} \mathbf{W}^{(l)}  \mathbf{h}_j^{(l-1)},\\
&\mathbf{h}^{(l)}_i = \sigma \big(\m^{(l)}_i + \alpha_{ii}^{(l)} \mathbf{W}^{(l)} \mathbf{h}^{(l-1)}_i \big),
\end{aligned}
\ee
where $\sigma (\cdot)$ is a non-linear activation function.

\subsection{Directional Aggregation Operators}
\label{subsec:dgn}
Directional Graph Network (DGN) \cite{beaini2021directional} is proposed to distinguish messages from neighbours with different topological importance.
%for the $\agg$ operation in Eq.~\eqref{def:message}.
%\sitao{matrices or operators?}, 
DGN explicitly assigns different weights to nodes within a neighbourhood directional using two aggregation matrix, 
namely the directional average matrix $\mathbf{B}_{\text{av}}$ and the directional derivative matrix $\mathbf{B}_{\text{dx}}$ defined as follows:
\footnote{In practice, row-normalized $\widetilde{\nabla \boldsymbol{\phi}}$ is used instead where $\widetilde{\nabla \boldsymbol{\phi}}_{i, :}=\frac{\mathbf{\nabla \boldsymbol{\phi}}_{i, :}}{\| \mathbf{\nabla \boldsymbol{\phi}}_{i, :} \|_{1} + \epsilon_0}$ with a small positive number $\epsilon_0$.}:
\be
\label{eq:directedagg}
% & \mathbf{B}_{\mathrm{av}}(\nabla \boldsymbol{\phi})_{i, :} =  |\widetilde{\nabla \boldsymbol{\phi}}_{i, :}|, \ \ \ \mathbf{B}_{\mathrm{dx}}(\nabla \boldsymbol{\phi})_{i, :} = \widetilde{\nabla \boldsymbol{\phi}}_{i, :} - \text{diag}(\widetilde{\nabla \boldsymbol{\phi}} \mathbf{1})_{i, :}. 
% & \widetilde{\nabla \boldsymbol{\phi}}_{i, :} = \left\{\begin{matrix}
% \frac{\mathbf{\nabla \boldsymbol{\phi}}_{i, :}}{\| \mathbf{\nabla \boldsymbol{\phi}}_{i, :} \|_{1}}, & \mathbf{\nabla \boldsymbol{\phi}}_{i, :} \neq \mathbf{0} \\
% \mathbf{0}, & \text{otherwise}
% \end{matrix}\right.,
\mathbf{B}_{\mathrm{av}}(\boldsymbol{\phi}) \!=\!  |\nabla \boldsymbol{\phi}|, \; 
\mathbf{B}_{\mathrm{dx}}(\boldsymbol{\phi}) \!=\! \nabla \boldsymbol{\phi} - \text{diag}(\nabla \boldsymbol{\phi} \mathbf{1}),
\ee
% \red{Can the argument $\nabla \boldsymbol{\phi}$ be removed 
% from $\mathbf{B}_{\mathrm{av}}(\nabla \boldsymbol{\phi})$
% and $\mathbf{B}_{\mathrm{dx}}(\nabla \boldsymbol{\phi})$ to make notation simpler?}
% \blue{If we remove it, how to show that later on we use another $B_{av}$ and $B_{dx}$ based on $\phib^{(1)}(\gamma)$ from $\mathbf{L}^{(1, \gamma)}$}
% \red{If a comparison is needed later, then replace
% $\mathbf{B}_{\mathrm{av}}(\nabla \boldsymbol{\phi})$ by
% $\mathbf{B}_{\mathrm{av}}(\boldsymbol{\phi})$.
% If no comparison is needed, we can say we take 
% $\boldsymbol{\phi}=\phib^{(1)}(\gamma)$ in (6).}
% \blue{OK, I will change the later on notation according to this}
where $\nabla \boldsymbol{\phi}=(\nabla \phi_{ij})\in \Rbb^{N\times N}$ is the graph vector field of a graph signal $\boldsymbol{\phi}=(\phi_i) \in \Rbb^{N}$ defined as follows:
\be
\label{eq:vectorfield}
\nabla\phi_{ij} = \begin{cases}
\phi_j - \phi_i, & e_{ij} \in \mathcal{E} \\
0, & \text{otherwise}
\end{cases}.
\ee
%(see \ref{def:diffusion-dist}) 

%indicating the direction of message passing on graph. 
%When used for aggregation, $\mathbf{B}_{\text{av}}$ and $\mathbf{B}_{\text{dx}}$ explicitly assign different importance weights to nodes within a neighbourhood (while GAT achieves this implicitly with learnable weights). 
In \cite{beaini2021directional}, $\boldsymbol{\phi}$
is taken to be $\boldsymbol{\phi}^{(1)}$,
the eigenevector corresponding to the smallest positive
eigenvalue of $\mathbf{L}_{\rw}$.
Note that $\nabla \boldsymbol{\phi}^{(1)}$
is the dominant direction of the graph diffusion process \cite{nadler2005diffusion}.
For message passing,
DGN regards the $\mathbf{B}_{\mathrm{av}}$ and $\mathbf{B}_{\mathrm{dx}}$ as edge weights for aggregation:
\be
\begin{aligned}
\label{eq:dgn}
&\resizebox{1\hsize}{!}{$\m^{(l)}_i = \sum_{j: v_j \in \{ \mathcal{N}(v_i)\}}  \mathbf{W}^{(l)}\big[\big(\mathbf{B}_{\mathrm{av}}(\boldsymbol{\phi}^{(1)})\big)_{ij} \mathbf{h}_j^{(l-1)} || \big(\mathbf{B}_{\mathrm{dx}}(\boldsymbol{\phi}^{(1)})\big)_{ij} \mathbf{h}_j^{(l-1)}\big],$} \\
&\mathbf{h}^{(l)}_i = \text{MLP}\left([\m^{(l)}_i ||  \mathbf{h}^{(l-1)}_i]\right). 
\end{aligned}
\ee
% With the directional aggregation defined by $\nabla \boldsymbol{\phi}^{(k)}$,
% the graph message passing becomes topology-aware as it follows the dominant directions of the graph diffusion process \cite{nadler2005diffusion}. Beaini \etal{} \cite{beaini2021directional} treat the type of Laplacian matrix (the choices being $\mathbf{L}$, $\mathbf{L}_{\mathrm{rw}}$ and $\mathbf{L}_\mathrm{sym}$) as a hyperparameter and tune it for each individual benchmark 
%\sitao{How does the authors $\mathbf{B}_{\text{av}}$ and $\mathbf{B}_{\text{dx}}$? Use it to replace the aggregation matrix?}.
% Though not mentioned explicitly, the empirical results given in 
% \cite{dgn2020repo} showed that using $\mathbf{L}$
% has better performance than using $\mathbf{L}_{\mathrm{rw}}$ and $\mathbf{L}_\mathrm{sym}$. 
%\sitao{This matrix can reduce diffusion distance and capture long-range information?}
The authors in \cite{beaini2021directional} claim that $\mathbf{B}_{\mathrm{av}}$ and $\mathbf{B}_{\mathrm{dx}}$ indicate a direction that can reduce diffusion distance. %which will be introduced in the following subSection  \sitao{Check if this sentence is correct}
And we refer to the operation $\mathbf{B}_{\{\mathrm{av}, \mathrm{dx}\}} (\boldsymbol{\phi}^{(1)}) \mathbf{H}$ as the directional aggregation.

\subsection{Spectral-based Nodes Relative Distances}
The relative position of nodes on the graph could be told by eigenvectors of Laplacian matrices \cite{kreuzer2021rethinking}. As the graph data is non-Euclidean, various measurements are proposed to describe the distance between nodes \cite{nadler2005diffusion, Qiu2007, Belkin2003}.
Among them,
the diffusion distance is often used to model how node $v_i$ influence node $v_j$  by considering random walks along edges \cite{beaini2021directional}.
At the initial step,
the random walk start at a node, 
and moves to one of its neighbours at the next step.
The diffusion distance between $v_i$ and $v_j$ is proportional to the probability that the random walk starting at node $v_i$ meets the random walk starting at node $v_j$ at step $t$.
Based on $\mathbf{L}_\rw$,
the \textbf{diffusion distance} is calculated as (see \cite{Coifman2006Diffusion}):
\be
\label{def:diffusion-dist}
d_t(v_i, v_j; \mathbf{L}_\rw) = \Big( \sum_{k=1}^{n-1} e^{-2t \lambda^{(k)}}(\phi_i^{(k)} - \phi_j^{(k)})^2 \Big)^\frac{1}{2}.
\ee
Another measure of the distance between $v_i$ and $v_j$ 
is the \textbf{spectral distance} defined as (see \cite{Belkin2003}):
\be
\label{def:spectral-dist}
d_{s}(v_i, v_j; \mathbf{L}_\rw) = | \phi_i^{(1)}- \phi_j^{(1)}|.
\ee
Note that $\phib^{(1)}$ gives positional information of nodes \cite{dwivedi2020generalization, kreuzer2021rethinking, muller2023attending}
and $\B_{\mathrm{av}}(\phib^{(1)})= (d_{s}(v_i, v_j; \mathbf{L}_\rw))$.
A more general definition of $d_s$ involves the $k$ eigenvectors
corresponding to the $k$ smallest eigenvalues,
but here we take $k=1$ for simplicity and informativeness.  

\subsection{Homophily Metrics}
% Homophily and heterophily are both properties of of a graph $G := (\mathcal{V}, \mathcal{E})$.
In the context of graph representation leanring, homophily refers to the tendency for nodes in a graph to share the same labels with their neighbours \cite{mcpherson2001birds}. Heterophily, conversely, describes the tendency for nodes to connect with other nodes with different labels. In this study, we employ the node homophily metric $H_{\mathrm{node}} \in [0, 1]$ defined as (see \cite{pei2020geom} \footnote{See more metrics in \cite{luan2023graph}.}):
\be
    \label{eq:node-homo}
    H_{\mathrm{node}} =
    \frac{1}{|\mathcal{V}|} \sum_{u \in \mathcal{V}}\frac{|y_u = y_v: v \in \mathcal{N}(v)| }{|\mathcal{N}(u)|}.
\ee
Graphs with strong homophily have large $H_{\mathrm{node}}$ (typically between 0.5 and 1);
on the contrary, heterophilic graphs have small $H_{\mathrm{node}}$ (typically $ < 0.5$).
Various alternative metrics including edge homophily \cite{zhu2020beyond}, 
class homophily \cite{lim2021new}, aggregation homophily \cite{luan2022revisiting}, graph smoothness value \cite{luan2022we} and classifier-based performance metrics \cite{luan2023graph} can also be used in practice.

\section{The Directional Attention Mechanism}
\label{sec:DAM}

In this section, we propose Directional Graph Attention Network (DGAT), 
which aims to enhance  the graph attention mechanism to capture long-range neighborhood information on graphs with different homophily levels, especially on heterophilic graphs. In Section~\ref{sec:para-lap}, we first propose a new parameterized normalized Laplacian matrix, which defines a new class of Graph Laplacian with more general directional aggregation. We then elaborate on the global directional flow in DGAT, which is defined based on the vector field utilizing the low-frequency eigenvector of this new class of Laplacian matrices. More specifically, two new mechanisms are introduced: the topology-guided neighbour rewiring (Section~\ref{sec:neighbour} and Section~\ref{sec:neighbour2}) and the global directional attention (Section~\ref{sec:global-direction}). Finally, we summarize the architecture of DGAT in Section~\ref{sec:dgat-arch-summary}.

\subsection{Parameterized Normalized Laplacian}
\label{sec:para-lap}
In order to gain a more refined control over the directional aggregation, we define a new class of Laplacian matrices.
\begin{definition}
\label{def:aug-lap}
A parameterized normalized Laplacian matrix is defined as
\be \label{eq:plaplacian}
\mathbf{L}^{(\alpha,\gamma)} 
= \gamma [\gamma \D + (1-\gamma )\I]^{-\alpha} \mathbf{L}
[\gamma \D + (1-\gamma )\I]^{\alpha-1}
\ee
and the corresponding parameterized normalized adjacent matrix  
is defined as
\be \label{def:adj}
\P^{(\alpha,\gamma)} =  \I-\mathbf{L}^{(\alpha,\gamma)},
\ee
where the parameters $\gamma \in (0, 1]$ and $\alpha \in [0, 1]$. 
\end{definition}

% new class of normalized Laplacian matrix (as defined in \eqref{eq:plaplacian}) 
% The two parameters $\gamma$ and $\alpha$ control the eigenvalues and eigenvectors of $\mathbf{L}^{(\alpha,\gamma)}$. 
%This new class of Laplacian matrix also help us to extend Theorem \ref{thm:gradient-original} 
%\sitao{Why do you suddenly mention Theorem \ref{thm:gradient-original}? What is the relation? Is this theorem a new one or from literature?}.
The normalized Laplacian matrices defined in Eq.~\eqref{def:lap} are special cases of this new class of Laplacian matrices: when $\alpha=1$ and $\gamma=1$, $\mathbf{L}^{(\alpha,\gamma)}=\mathbf{L}_\rw$; when $\alpha=\frac{1}{2}$ and $\gamma=1$, $\mathbf{L}^{(\alpha,\gamma)}=\mathbf{L}_\sym$.  
Although we cannot choose $\alpha$ and $\gamma$ such that 
$\mathbf{L}^{(\alpha,\gamma)}$ becomes $\mathbf{L}$, %we have the following result:
% \be 
% \label{proof:inf}
% \lim_{\gamma \to 0} \frac{1}{\gamma}\mathbf{L}^{(\alpha, \gamma)}
% &= \lim_{\gamma \to 0} (\gamma \D + (1-\gamma )\I)^{-\alpha}
% &\mathbf{L} \times \lim_{\gamma \to 0} 
% (\gamma \D + (1-\gamma )\I)^{\alpha-1} 
%       = \mathbf{L} 
% \ee
we have $\lim_{\gamma \to 0} \frac{1}{\gamma}\mathbf{L}^{(\alpha, \gamma)} = \mathbf{L}$,
which implies that when $\gamma$ is small enough, 
an eigenvector of $\mathbf{L}^{(\alpha,\gamma)}$ is a good approximation to
an eigenvector of $\mathbf{L}$.

% The adjacency matrix is defined accordingly as follows. 
% \begin{definition}
% \label{parameterized_matrix}
% \label{def:aug-rw}
% The parameterized normalized adjacent matrix corresponding to 
% $\mathbf{L}^{(\alpha,\gamma)}$ is defined as
% $$
% \P^{(\alpha,\gamma)} \coloneqq  \I-\mathbf{L}^{(\alpha,\gamma)}
% $$
% where the parameters $\gamma \in (0, 1], \alpha \in [0, 1]$.
% \end{definition}

The following theorem justifies the definition of $\P^{(1,\gamma)}$ as a random walk matrix.
\begin{theorem}
\label{thm:parameterized-matrix}
The $\P^{(\alpha, \gamma)}$ defined in \eqref{def:adj} is non-negative (\ie{} all of its elements are non-negative), and when $\alpha=1$, $\P^{(\alpha, \gamma)}\1= \1$. 
See the proof in Appendix \ref{appendix:proof_parameterized_matrix}.
\end{theorem}

The theorem indicates $\P^{(1, \gamma)}$ is a random walk matrix,
which is a generalization of the classic random walk matrix $\D^{-1}\A$.
The properties of eigenvalues of $\mathbf{L}^{(\alpha,\gamma)}$ 
are given in the following Theorem.

\begin{theorem}
\label{thm:eigenplaplaican}
    Suppose the graph $\mathcal{G}$ is connected.
    Then the symmetric $\mathbf{L}^{(1/2,\gamma)}\in \Rbb^{n\times n}$ has the eigendecomposition:
    \be \label{eq:gled}
    \mathbf{L}^{(1/2,\gamma)} = \U \bLambda^{(\gamma)} \U^\top,
    \ee
    where $\U\in \Rbb^{N\times N}$ is orthogonal and $\bLambda^{(\gamma)}=\diag(\lambda^{(i)}(\gamma))$,
    \be \label{eq:glevrange}
    0 =\lambda^{(0)}(\gamma) < \lambda^{(1)}(\gamma)
    \leq \cdots \leq \lambda^{(N-1)}(\gamma) \leq 2.
    \ee
%$\lambda^{(1)}(\gamma)\neq 0$, and
Each $\lambda^{(i)}(\gamma)$ is strictly increasing with respect to $\gamma$ for $i=1:N-1$.
Furthermore, $\mathbf{L}^{(\alpha,\gamma)}$ has the eigendecomposition
\be \label{eq:gled-alpha}
\resizebox{1\hsize}{!}{$ 
\mathbf{L}^{(\alpha,\gamma)} 
= \left([\gamma\D+(1-\gamma)\I]^{\frac{1}{2}-\alpha} \U \right) \bLambda^{(\gamma)} 
\left([\gamma\D+(1-\gamma)\I]^{\frac{1}{2}-\alpha}\U\right)^{-1} $},
\ee
\ie{} $\mathbf{L}^{(\alpha,\gamma)}$ share the same eigenvalues as $\mathbf{L}^{(1/2,\gamma)}$ and the columns of $[\gamma\D+(1-\gamma)\I]^{\frac{1}{2}-\alpha}\U$ are the corresponding eigenvectors. See proof in Appendix \ref{appendix:proof_eigenplaplaican}.
\end{theorem}

The following theorem shows the monotonicity of the diffusion distance with respect to the spectral distance.
Here both distances are defined in terms of eigenvectors of 
$\mathbf{L}^{(1,\gamma)}$,
cf.\ \eqref{def:diffusion-dist} and \eqref{def:spectral-dist},
which involve the eigenvectors of $\mathbf{L}_\rw$.
%we extend \cite{beaini2021directional} to broader settings as follows.

% \blue{Since gradient step is not required, we need to change the name of the theorem too.}
\begin{theorem} 
\label{thm:gradient-v2}
Let $v_i$, $v_j$ and $v_m$ be nodes of the graph $\mathcal{G}$ such that 
$d_{s}(v_m, v_j; \mathbf{L}^{(1,\gamma)}) < d_{s}(v_i, v_j; \mathbf{L}^{(1,\gamma)})$.
% \ie{}
% $$
% |\phi_m^{(1)}(\gamma) - \phi_j^{(1)}(\gamma)| < |\phi_i^{(1)}(\gamma) - \phi_j^{(1)}(\gamma)|,
% $$
% where $\phib^{(1)}(\gamma)$ is the eigenvector corresponding to the smallest non-trivial eigenvalue $\lambda^{(1)}(\gamma)$ of $\mathbf{L}^{(1, \gamma)}$. 
Then there is a constant $C$ such that for $t\geq C$,  
\be
    d_t(v_m, v_j; \mathbf{L}^{(1,\gamma)}) < d_t(v_i, v_j; \mathbf{L}^{(1,\gamma)}).
\ee
Furthermore, the ratio $d_t(v_m, v_j; \mathbf{L}^{(1,\gamma)})/d_t(v_i, v_j; \mathbf{L}^{(1,\gamma)})$
is proportional to $e^{-\lambda^{(1)}(\gamma)}$. 
\end{theorem}
 %\sitao{This means $m$ has more impact on $j$ than $i$, so we need to connect $m$ and $j$? Do you have different usage compared with \cite{beaini2021directional}, \eg{}  \cite{beaini2021directional} doesn't connect new nodes but you do?}
 %\blue{DGN use this to show that Bav and Bdx (which is defined according to $\phi_i - \phi_j$) provide directional information and then use them as aggregation matrices.}

Theorem \ref{thm:gradient-v2}\footnote{Theorem \ref{thm:gradient-v2} substantially extends the scope of \cite{beaini2021directional}, which deals with the diffusion distance defined by the random walk Laplacian $\mathbf{L}_{\rw}$. Furthermore, Theorem \ref{thm:gradient-v2} overcomes some shortcomings of \cite{beaini2021directional}, see details in Appendix \ref{appendix:gradient-v2}.} shows that the spectral distance can be used as a good indicator of the diffusion distance, \ie{} if node $i$ has a larger spectral distance to node $j$ than node $m$, then node $i$ would also have a larger diffusion distance to node $j$ after enough time steps. According to \eqref{def:diffusion-dist}, we needs to find all eigenvalues and eigenvectors of the Laplacian to calculate the diffusion distance, which is computationally expensive. But as Theorem \ref{thm:gradient-v2} shows, we can easily compute the spectral distance and use it as a surrogate function of diffusion distance.
%\sitao{Check if this rewritten sentence is correct}
% It is worth to note that Theorem \ref{thm:gradient-v2} cannot be used to compare the diffusion distance between arbitrary pairs of nodes.
% \red{What do you mean? Do you mean $v_j$ needs to be in both distances?}
% \blue{Yes, we need the common node $v_j$, maybe it does not matter, will double check this}

% As $\gamma \in (0, 1] \nearrow$,
% $\lambda^{(1)}(\gamma) \nearrow$,
% $\mathbf{L}^{(1,\gamma)} \longrightarrow = \mathbf{L}_{rw}$.
% As $\gamma \in (0, 1] \searrow$ with $\alpha = 1$,
% $\lambda^{(1)}(\gamma) \longrightarrow  \lambda^{(1)}$.

% \red{homophilic graph or homophily graph? Be consistent.
% Check all places}

For graphs with different homophily levels, the node would prefer neighborhood information from different diffusion distances. Specifically, \textbf{nodes with small diffusion distances would be favored on homophilic graphs and nodes with large diffusion distances would be helpful on heterophilic graphs to reduce the effect of noises \cite{topping2021understanding}.}
%\sitao{ It's better to find reference to justify the relation between the diffusion distance and noise.}.
%\blue{I think we just mention it here that why the proposed L is useful, and leave the discussion about heterophily to other sections}
Adjusting $\alpha$ and $\gamma$  can change the eigenvalues and eigenvectors of  $\mathbf{L}^{(\alpha,\gamma)}$, leading to adjustment of diffusion distances. For instance, we can employ a smaller $\gamma$ on heterophilic graphs to help the propagation of long-range neighborhood information and employ a larger $\gamma$ on homophilic graphs to capture the useful local information. Therefore, we can tune $\alpha$ and $\gamma$ as hyperparameters to control the message passing in a nuanced way.
%\red{Does the last part of Thm 3.3 provide some guidance
% on choosing $\gamma$ (i.e., large or small)? 
% If so, state it here.}
% In the next section, we provide qualitative guidance on choosing $\gamma$ and $\alpha$ for different input graphs based on their heterophily level.

\subsection{Topology-Guided Neighbour Pruning}
\label{sec:neighbour}
% Neighbour pruning is one of the two mechanisms that implement the global directional flow in our model. 
%The intuition behind this mechanism is that we want 
In order to filter out the noise, \ie inter-class information, during feature aggregation, we propose a novel topology-guided neighbour pruning strategy to remove the noisy neighbors.
% Unlike existing rewiring strategies \cite{},
% Our strategy is more computationally efficient with the cost of .
% \blue{I think other existing rewiring strategy cost much more than ours in my memory, but i need to find some references.}
% described as follows.
% The discussion about the node distance is based on the proposed $\mathbf{L}^{(1,\gamma)}$ and the $\mathbf{L}^{(1,\gamma)}$ is omitted in notation for simplicity.

The diffusion distance can be regarded as a measure of node dependency \cite{coifman2005geometric} or a criterion for the efficiency of message passing between nodes that are not directly connected. For connected nodes $v_i$ and $v_j$,
although $d_t(v_i, v_j; \mathbf{L}^{(1,\gamma)})$ may be large, the message from $v_j$ still reaches $v_i$ in one step via $e_{ij}$ in GNNs' message passing.

Based on Theorem \ref{thm:gradient-v2}, we propose to use $d_s(v_i, v_j; \mathbf{L}^{(1,\gamma)})$ to determine whether the $v_j \in \calN(v_i)$ tends to propagate noisy information to $v_i$ under the homophily and heterophily assumptions, respectively.

Let $\phib^{(1)}$ be the eigenvector of $\mathbf{L}^{(1,\gamma)})$
corresponding to the smallest non-trivial eigenvalue. For node $i$ and its neighbor $v_j \in \calN(v_i)$, assume $\phi_j^{(1)}(\gamma) > \phi_i^{(1)}(\gamma)$. For $v_k \notin \calN{(v_i)}$ satisfying the following conditions
\be
\begin{aligned}
\label{eq:vk_condition}
\phi_k^{(1)}(\gamma) > \phi_i^{(1)}(\gamma), d_{t}(v_k, v_i; \mathbf{L}^{(1,\gamma)}) > \frac{1}{2} 
d_{t}(v_i, v_j; \mathbf{L}^{(1,\gamma)}),
\end{aligned}
\ee
we have $d_{t}(v_k, v_j; \mathbf{L}^{(1,\gamma)}) < d_{t}(v_k, v_i; \mathbf{L}^{(1,\gamma)})$.
This means node $v_j$ (which is connected to node $v_i$) with smaller diffusion distance are more likely to be used to accelerate message passing from node $v_k$ to node $v_i$. %\sitao{Why? Does $k$ need to connect $j$?}.
Based on the above discussion about the condition of $v_k$ should satisfy,
we propose the following neighbour pruning strategy to remove neighbors that will carry more higher-order neighborhood information in homophilic graphs and eliminate neighbors which can carry less higher-order information in heterophilic graphs.
With a pre-defined threshold $\epsilon$\,\footnote{We treat $\epsilon$ as a hyperparameter, and it is tuned independently for different datasets},
we compare each $d_s(v_i, v_j; \mathbf{L}^{(1,\gamma)})$ against $\epsilon$ and decide whether $e_{ij}$ should be pruned regarding the following two cases.

\paragraph{The homophily case}
It follows that the smaller $d_{s}(v_i, v_j; \mathbf{L}^{(1,\gamma)})$ is,
the larger the set of nodes satisfying the condition in \eqref{eq:vk_condition} is.
%range of $d_{t}(v_k, v_i; \mathbf{L}^{(1,\gamma)})$.
Which means that neighbour $v_j$ with smaller $d_{s}(v_i, v_j; \mathbf{L}^{(1,\gamma)})$ tends to carry more diverse information propagated from nodes with wider range of diffusion distance to $v_i$.
Under the homophily assumption,
% we assume such $v_j$ shares the same label with $v_i$.
nodes are more likely to share labels with local neighbours than higher-order neighbors.
Thus, GNNs should receive more information from local neighbours $\calN(v_i)$ with less potential to carry higher-order neighborhood information. Therefore,  edges with $d_{s}(v_i, v_j; \mathbf{L}^{(1,\gamma)}) < \epsilon$ are pruned, as they introduce more noise which may prohibit graph representation learning on a homophily graph.

%$|\phi_k^{(1)}(\gamma) - \phi_j^{(1)}(\gamma)| < |\phi_i^{(1)}(\gamma) - \phi_j^{(1)}(\gamma)|$.

\paragraph{The heterophily case}
Conversely, the larger $d_{s}(v_i, v_j; \mathbf{L}^{(1,\gamma)})$ is, the more likely that neighbor $v_j$ 
%Which means that neighbour $v_j$ with larger $d_{s}(v_i, v_j; \mathbf{L}^{(1,\gamma)})$ t
tends to carry less higher-order information to $v_i$. Thus edges with $d_{s}(v_i, v_j; \mathbf{L}^{(1,\gamma)}) > \epsilon$ are pruned.

\begin{figure*}[h!]
    \centering
    \resizebox{1\hsize}{!}{
    \includegraphics[scale=0.15]{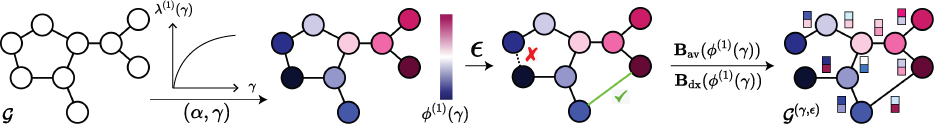}}
    \caption{The workflow of DGAT.}  
    \label{fig:DGAT}
\end{figure*}

\subsection{Topology-Guided Edge Adding}
\label{sec:neighbour2}
For heterophilic graphs, edges pruning is not enough. Unlike the neighbor denoising for the heterophily case in Section~\ref{sec:neighbour}, where the edges are assumed to be inter-class, we are going to add intra-class edges to provide extra beneficial information for message propagation under the assumption that disconnected nodes with large diffusion distance are more likely to share labels in a heterophily graph \cite{topping2021understanding}. However, it tends to bring a large amount of unhelpful edges if we connect every pair of nodes with large diffusion distance.
%but adhere to the graph rewiring principle that the rewiring should be surgical \cite{}.
We propose the following strategy to identify the most helpful nodes to add edges on.

Let $\phi_m^{(1)}(\gamma) = \min_{k} \phi_k^{(1)}(\gamma)$, $\phi_n^{(1)}(\gamma) = \max_{k} \phi_k^{(1)}(\gamma)$ and $M = \frac{1}{2} (\phi_m^{(1)}(\gamma) + \phi_n^{(1)}(\gamma))$. For each node $v_i$ with $\phi_i^{(1)}(\gamma) < M$, we connect this $v_i$ to $v_n$. The newly added edge helps propagating information from any unconnected nodes with a large diffusion distance to $v_i$. To explain it, consider $v_j$ with the property
\be
v_j \notin \calN(v_i), \ \ d_{s}(v_i, v_j; \mathbf{L}^{(1,\gamma)}) > M,
\ee
we have $d_{t}(v_n, v_j; \mathbf{L}^{(1,\gamma)}) < d_{t}(v_n, v_i; \mathbf{L}^{(1,\gamma)})$.

Then the newly added edge $e_{in}$ makes $v_n$ a useful neighbor of $v_i$,
which promotes message passing from all $v_j$ with a large diffusion distance ($> M$) to $v_i$.

For each node $v_i$ with $\phi_i^{(1)}(\gamma) > M$,
we connect node $v_i$ and node $v_m$
and the reason is similar to that given above.

\subsection{Global Directional Attention}
\label{sec:global-direction}
% % As mentioned in the preceding section, 
The original GAT employs multi-head attention to enrich the model's capabilities and stabilize the learning process \cite{velivckovic2017graph},
% %In specific, $M$ independent attention mechanisms are executed and combined.
% %more specifically, 
but each attention head is feature-based
and does not adapt to the specific characteristics of the global topology. 
In contrast,
the DGN projects incoming messages into directions defined by the graph topology \cite{beaini2021directional}
but utilizes fixed weights within node neighbourhoods,
consequently losing flexibility in local aggregation.
To remedy these shortcomings, 
we propose a global directional attention mechanism, 
which is characterized by spectral-based edge features and topology-aware attention.

\subsubsection{Spectral-based edge features}
\label{sec:edgefeature}
We propose to define the feature of $e_{ij} \in \mathcal{E}$
as
\be
\label{eq:edgefeature_dgn}
\f^{(i,j)} = \big[ \mathbf{B}_{\mathrm{av}}(\boldsymbol{\phi}^{(1)}(\gamma))(i,j), \mathbf{B}_{\mathrm{dx}}(\boldsymbol{\phi}^{(1)}(\gamma))(i,j) \big]^\top,
\ee
where 
$\mathbf{B}_{\mathrm{av}}(\boldsymbol{\phi}^{(1)}(\gamma))$ and $\mathbf{B}_{\mathrm{dx}}(\boldsymbol{\phi}^{(1)}(\gamma))$ are matrices corresponding to the aggregation and diversification operations utilizing the vector field $\nabla \boldsymbol{\phi}^{(1)}(\gamma)$
(see \eqref{eq:directedagg}).
By Theorem \ref{thm:gradient-v2}, $\mathbf{B}_{\{ \mathrm{av}, \mathrm{dx} \}}(\boldsymbol{\phi}^{(1)}(\gamma))(i,j)$
contains information of the diffusion distance between $v_i$ and $v_j$,
which describes the relative position of $v_i$ and $v_j$ on the graph.
Instead of regarding them as edge weights,
we use this node relative distance information to assist the learning of attention between nodes.
%\red{This is not clear to me. But you don't need to make a change if it's clear to you.}

\subsubsection{Topology-aware attention}
\label{sec:global_att}
We propose to define the attention score between $v_i$ and $v_j$ before normalization as:
\be
\label{eq:dgat_attention}
\resizebox{1\hsize}{!}{$r_{ij}^{(l)} = \text{LeakyReLU}((\mathbf{a}^{(l)})^{\top} [\mathbf{W}_n^{(l)} \mathbf{h}_i^{(l-1)} || \mathbf{W}_n^{(l)} \mathbf{h}_j^{(l-1)} || \mathbf{W}_e^{(l)} \f^{(i, j)}]).$},
\ee
where %$\f^{(i, j)}$ is the edge feature of $e_{ij}$ in \eqref{eq:edgefeature_dgn},
$\mathbf{W}_n^{(l)}$ and $\mathbf{W}_e^{(l)}$ are learnable weights for node representation and edge features respectively.
The normalized attention score is calculated according to \eqref{eq:attentionscore}.
Compared with GAT,
the additional number of parameters in \eqref{eq:dgat_attention} is negligible,
since $\f^{(i, j)}\in \Rbb^2$.
The normalized attention score $( \alpha_{ij}^{(l)})$ is calculated according to graph connectivity by \eqref{eq:attentionscore}.

\iffalse
\subsection{Directional Disentangled Graph Attention Mechanism}

To model the diversity of edge directionality, we propose Directional Graph Attention Mechanism to disentangle the learning process of the edge weights for each edge. Specifically, the pair of nodes $(i,j)$, we propose to learn $e_{i \rightarrow j}$ and $e_{j \rightarrow i}$ with two independent sets of weights.
\fi

\subsection{Directional Graph Attention Network}
\label{sec:dgat-arch-summary}
Figure \ref{fig:DGAT} %\red{Check this} \blue{I am working on this} 
provides a schematic representation of the workflow of DGAT model. Here we use $\alpha = 1$ as an example. Given the graph $\mathcal{G} = (\mathcal{V}, \mathcal{E})$ and the hyperparameter $\gamma$, the DGAT calculates the first non-trivial eigenvector $\boldsymbol{\phi}^{(1)}(\gamma)$ of $\mathbf{L}^{(1, \gamma)}$.
Based on $\nabla \boldsymbol{\phi}^{(1)}(\gamma)$ and the pruning threshold $\epsilon$,
the neighbour pruning or the edge adding strategy is performed,
and is tailored to either a homophily or heterophily scenario.
We denote the rewired graph as $\mathcal{G}^{(\gamma, \epsilon)} = (\mathcal{V}, \mathcal{E}^{(\gamma, \epsilon)})$ with self-loop added to it
%and the neighbourhood of $v_i$ in $\mathcal{G}^{(\gamma, \epsilon)}$ as $\calN^{(\gamma, \epsilon)}(v_i)$
\footnote{Although the edge adding strategy is hyperparameter-free, the $\epsilon$ is still there to distinguish the original graph and the rewired graph.}.
Edge features are defined according to \eqref{eq:edgefeature_dgn} based on the pruned graph.
Utilizing multi-heads attention \cite{vaswani2017attention, velivckovic2017graph},
the $l^{th}$ layer of DGAT updates the representation of $v_i$ as
\be
\label{eq:dgat_multihead}
\resizebox{1\hsize}{!}{$
\h_i^{(l)} =\left( \bigg\Arrowvert^{M}_{m=1} \sigma \big( 
\sum_{j: e_{ij} \in  \mathcal{E}^{(\gamma, \epsilon)}} \alpha_{ij}^{(l, m, \gamma, \epsilon)} \W^{(l, m)} \h_j^{(l-1)}
\big)   \right)\W^{(l)},$}
\ee
where $M$ is the number of heads,
$\alpha_{ij}^{(l, m, \gamma, \epsilon)}$ is the topology-aware attention on $m^{th}$ head defined in Section~\ref{sec:global_att}.

\begin{table*}[htbp]
   \resizebox{1\hsize}{!}{
  \begin{tabular}{ccccccccccc}
  \toprule
    $\mathbf{L}^{(\alpha, \gamma)}$&
    $\text{ATT}_\mathbf{e}$&
    $\mathcal{R}(\mathcal{G})$&
    roman-empire&
    amazon-ratings&
    minesweeper&
    tolokers&
    questions&
    squirrel filtered&
    chameleon filtered&
    \\
    \midrule
    \multirow{1}[0]{*}
    &
    &
    &
    80.87 $\pm$ 0.30&
    49.09 $\pm$ 0.63&
    92.01 $\pm$ 0.68&
    83.70 $\pm$ 0.47&
    77.43 $\pm$ 1.20&
    35.62 $\pm$ 2.06&
    39.21 $\pm$ 3.08&
    \\
    \midrule
    \multirow{1}[0]{*}
    &
    &
    $\checkmark$ &
    81.21 $\pm$ 0.71&
    47.27 $\pm$ 0.52&
    92.86 $\pm$ 0.53&
    84.28 $\pm$ 0.68&
    78.79 $\pm$ 1.01&
    40.45 $\pm$ 1.36&
    43.51 $\pm$ 5.06&
    \\
    \multirow{1}[0]{*}
    &
    $\checkmark$ &
    &
    82.50 $\pm$ 0.60&
    49.29 $\pm$ 0.57&
    92.07 $\pm$ 0.69&
    84.11 $\pm$ 0.41&
    OOM &
    36.50 $\pm$ 1.37&
    41.84 $\pm$ 2.76&
    \\
   
    $\checkmark$ &
        &
    $\checkmark$ &
    82.27 $\pm$ 0.60&
    47.69 $\pm$ 0.64&
    92.94 $\pm$ 0.55&
    84.59 $\pm$ 0.53&
    78.97 $\pm$ 1.04&
    41.49 $\pm$ 2.72&
    43.89 $\pm$ 4.67&
    \\
    \multirow{1}[0]{*}
       &
    $\checkmark$ &
    $\checkmark$ &
    86.82 $\pm$ 0.60&
    47.81 $\pm$ 0.54&
    93.15 $\pm$ 0.80&
    84.39 $\pm$ 0.49&
    OOM &
    41.48 $\pm$ 2.34&
    43.09 $\pm$ 4.13&
    \\
   
    $\checkmark$ &
    $\checkmark$ &
      &
    83.01 $\pm$ 0.52&
    \textbf{49.69 $\pm$ 0.51}&
    92.15 $\pm$ 0.71&
    84.11 $\pm$ 0.41&
    78.66 $\pm$ 0.97&
    37.83 $\pm$ 1.54&
    43.37 $\pm$ 3.01&
    \\
    \midrule
   
    $\checkmark$ &
    $\checkmark$ &
    $\checkmark$ &
    \textbf{87.27 $\pm$ 0.64}&
    48.03 $\pm$ 0.58&
    \textbf{93.27 $\pm$ 0.56}&
    \textbf{84.74 $\pm$ 0.59}&
    \textbf{79.55 $\pm$ 0.81}&
    \textbf{42.09 $\pm$ 2.65}&
    \textbf{44.16 $\pm$ 4.20}&
    \\
    \bottomrule
  \end{tabular}
  }
\caption{Ablation study on heterophily datasets proposed by \cite{platonov2023critical}.
The checkmark on $\mathbf{L}^{(\alpha, \gamma)}$ stands for adopting parameterized normalized Laplacian with optimal $\alpha$ and $\gamma$,
while unchecking $\mathbf{L}^{(\alpha, \gamma)}$ indicates that the default $\mathbf{L}^{(1, 1)} = \mathbf{L}_{\rw}$ is used.
The $\text{ATT}_\mathbf{e}$ stands for the global directional attention mechanism.
And $\mathcal{R}(\mathcal{G})$ stands for graph re-wiring strategy.
The OOM occurs for some specific choice of $\alpha$ and $\gamma$ in computing eigenvectors.
The best results are highlighted in \textbf{bold} format.}
\label{tab:heter_ablation}
\end{table*}

\section{Related Works}
\label{sec:related_works}
A general graph shift operator is proposed in \cite{eliasof2023improving} to include various kinds of adjacency and Laplacian matrices.
While \cite{eliasof2023improving} propose a learnable graph shift operation. 
The new class of Laplacians in this work distinguishes from them as its eigenvalues are bounded and offer a theoretical way to control the spectral property of graph. 

We have seen methods that enhance GNNs' performance by changing the graph structure \cite{rong2019dropedge, papp2021dropgnn}.
Graph rewiring,
including spatial rewiring \cite{abu2019mixhop} and spectral-based methods \cite{arnaiz2022diffwire},
are proposed to alleviate over-squashing \cite{alon2020bottleneck}.
While our method is characterized by its adaptiveness to different heterophily levels.

Several recent studies have incorporated the attention mechanism into GNNs \cite{Brody2021How, yun2019graph, hu2020heterogeneous, wang2020direct,luan2022revisiting},
and research interests in graph transformer are rising \cite{kreuzer2021rethinking, muller2023attending}
% The GATv2 \cite{Brody2021How} propose a dynamic feature-based graph attention variant,
% exhibiting superior performance compared to the original GAT.
In our implementation, 
we use the graph attention mechanism proposed in GAT \cite{velivckovic2017graph}.
Our proposed method is immune to the specific attention mechanism it can combine with, 
offering the potential to enhance various graph-based attention mechanisms including graph transformers.
% Since the method we proposed is immune to the type of attention mechanism it can combine with and can be used to enhance other attention mechanisms on graph.
Using the vanilla graph attention mechanism makes the effectiveness of DGAT more convincing,
as elaborated in Section~\ref{sec:synthetic}.

\section{Experiments}
\label{sec:experiments}
In this section, we evaluate the effectiveness of our proposed DGAT on synthetic and real-world benchmark datasets. In Section \ref{sec:ablation}, we conduct ablation experiments to  validate the effectiveness of each proposed component in DGAT. In Section \ref{sec:synthetic}, we generate synthetic graphs with various homophily levels and shows that DGAT can outperform the baseline models in each homophily level. In Section \ref{sec:benchmark}, we compare DGAT with several baseline and state-of-the-art (SOTA) models on 7 real-world benchmark datasets and the results show that DGAT outperform the SOTA models on 6 out of 7 node classification tasks.

\subsection{Ablation Study}
\label{sec:ablation}
This section conducts ablation studies to investigate the effectiveness of the usage of (1) parameterized normalized Laplacian $\mathbf{L}^{(\alpha, \gamma)}$, (2) the topology-guided graph rewiring strategy ($\mathcal{R}(\mathcal{G})$) and (3) the global directional attention ($\text{ATT}_\mathbf{e}$) in DGAT. %\sitao{Data split, hyperparameters, optimizer, ... etc.} 
The results on 7 heterophily datasets are summarized in Table \ref{tab:heter_ablation} \footnote{See Appendix \ref{app:ablation} for the ablation study on the other 9 small-scale datasets \cite{pei2020geom}.}.

It is observed from Table \ref{tab:heter_ablation} that: (1) even without leveraging $\mathbf{L}^{(\alpha, \gamma)}$, \ie{} we only use the default $\mathbf{L}_{\rw}$ with $\alpha=1$, $\gamma=1$, the topology-guided graph rewiring strategy and the global directional attention can bring improvement individually; (2) after incorporating with $\mathbf{L}^{(\alpha, \gamma)}$ \footnote{When $\mathbf{L}^{(\alpha, \gamma)}$ is utilized,
we optimize $\alpha$, $\gamma$ for each graph to find the most suitable relative position for nodes.}, the effectiveness of the above two components are further enhanced; (3) each component is indispensable for the success of DGAT and the full model performs the best on 6 out of 7 datasets (except for \textit{amazon-ratings}\footnote{It can be explained by the results in \cite{platonov2023critical}, which show that the GNNs yield negligible improvement over the graph-free models on \textit{amazon-ratings}.}) when we combine the 3 components together. %\sitao{It's better to give an explanation to why does the model cannot perform the best on amazon-ratings.}.

\begin{table}[htbp]
  \centering  
  \begin{tabular}{ccccc}
  \toprule
    $\mu$ 
    & GAT
    & GATv2
    & DGAT ($\gamma^{*}$)
    & DGAT ($\gamma^{-}$)
    \\
    \midrule
    0.0
    & 45.21
    & 48.66
    & \textbf{54.94 (0.1)}
    & 47.51 (0.7)
    \\

    0.1
    & 44.63
    & 50.65
    & \textbf{54.13 (0.3)}
    & 49.72 (0.9)
    \\
    
    0.2
    & 47.32
    & 54.46
    & \textbf{57.83 (0.3)}
    & 53.11 (1.0)
    \\
    
    0.3 
    & 56.59
    & 59.04
    &\textbf{64.33 (0.3)}
    & 59.37 (1.0) 
    \\
    
    0.4 
    & 63.21
    & 66.49
    & \textbf{73.54 (0.3)}
    & 66.71 (0.7)
    \\
    
    0.5
    & 74.14
    & 75.71
    &\textbf{82.95 (0.7)}
    & 78.81 (0.2) 
    \\
    
    0.6 
    & 83.96
    & 82.29
    & \textbf{90.56 (0.7)}
    & 88.30 (0.5)
    \\
    
    0.7
    & 89.33
    & 88.33
    & \textbf{96.00 (0.9)}
    & 94.85 (0.4)
    \\
    
    0.8
    & 94.92
    & 94.84
    & \textbf{98.67 (0.6)}
    & 97.84 (0.5)
    \\
    
    0.9
    & 98.62
    & 97.55
    & \textbf{99.58 (0.8)}
    & 98.21 (0.1)
    \\

  \bottomrule
  \end{tabular}

  \caption{Experimental results on synthetic graphs: average test accuracy (\%) with different homophily coefficients.
  The results are averaged over 5 random splits.
  The best results are in \textbf{bold} format. 
  For the DGAT model, we list the best and worst test runs (in the column DGAT ($\gamma^{*}$) and DGAT ($\gamma^{-}$) respectively) with different choices of  of $\gamma$ in the bracket, e.g. $54.94 (0.1)$ means that for $\gamma = 0.1$, the average test accuracy is $54.96\%$.}
  \label{table:syn_res}
\end{table}
\begin{table*}[h!]
 \resizebox{1\hsize}{!}{
  \begin{tabular}{lcccccccc}
  \toprule

    &
    roman-empire&
    amazon-ratings&
    minesweeper&
    tolokers&
    questions&
    squirrel-filtered&
    chameleon-filtered
    \\

    \midrule
    \multirow{1}[0]{*}
    GCN&
    73.69 $\pm$ 0.74& 
    48.70 $\pm$ 0.63&
    89.75 $\pm$ 0.52&
    83.64 $\pm$ 0.67&
    76.09 $\pm$ 1.27&
    39.47 $\pm$ 1.47&
    40.89 $\pm$ 4.12&
    \\
    
    %\midrule
    \multirow{1}[0]{*}
    SAGE&
    85.74 $\pm$ 0.67&
    \red{53.63 $\pm$ 0.39}&
    \textcolor{violet}{93.51 $\pm$ 0.57}&
    82.43 $\pm$ 0.44&
    76.44 $\pm$ 0.62&
    36.09 $\pm$ 1.99&
    37.77 $\pm$ 4.14&
    \\

    \multirow{1}[0]{*}
    GAT&
    80.87 $\pm$ 0.30&
    49.09 $\pm$ 0.63&
    92.01 $\pm$ 0.68&
    83.70 $\pm$ 0.47&
    77.43 $\pm$ 1.20&
    35.62 $\pm$ 2.06&
    39.21 $\pm$ 3.08&
    \\

    \multirow{1}[0]{*}
    GAT-sep&
    \blue{88.75 $\pm$ 0.41}&
    \textcolor{violet}{52.70 $\pm$ 0.62}&
    \blue{93.91 $\pm$ 0.35}&
    \textcolor{violet}{83.78 $\pm$ 0.43}&
    76.79 $\pm$ 0.71&
    35.46 $\pm$ 3.10&
    39.26 $\pm$ 2.50&
    \\

    \multirow{1}[0]{*}
    GT&
    86.51 $\pm$ 0.73&
    51.17 $\pm$ 0.66&
    91.85 $\pm$ 0.76&
    83.23 $\pm$ 0.64&
    77.95 $\pm$ 0.68&
    36.30 $\pm$ 1.98&
    38.87 $\pm$ 3.66&
    \\
    
    \multirow{1}[0]{*}
    GT-sep&
    87.32 $\pm$ 0.39&
    52.18 $\pm$ 0.80&
    92.29 $\pm$ 0.47&
    82.52 $\pm$ 0.92&
    78.05 $\pm$ 0.93&
    36.66 $\pm$ 1.63&
    40.31 $\pm$ 3.01&
    \\

    \midrule
    \multirow{1}[0]{*}
    H$_2$GCN&
    60.11 $\pm$ 0.52&
    36.47 $\pm$ 0.23&
    89.71 $\pm$ 0.31&
    73.35 $\pm$ 1.01&
    63.59 $\pm$ 1.46&
    35.10 $\pm$ 1.15&
    26.75 $\pm$ 3.64&
    \\
    
    \multirow{1}[0]{*}
    CPGNN&
    63.96 $\pm$ 0.62&
    39.79 $\pm$ 0.77&
    52.03 $\pm$ 5.46&
    73.36 $\pm$ 1.01&
    65.96 $\pm$ 1.95&
    30.04 $\pm$ 2.03&
    33.00 $\pm$ 3.15&
    \\

    \multirow{1}[0]{*}
    GPR-GNN&
    64.85 $\pm$ 0.27&
    44.88 $\pm$ 0.34&
    86.24 $\pm$ 0.61&
    72.94 $\pm$ 0.97&
    55.48 $\pm$ 0.91&
    38.95 $\pm$ 1.99&
    39.93 $\pm$ 3.30&
    \\

    \multirow{1}[0]{*}
    FSGNN&
    79.92 $\pm$ 0.56&
    \blue{52.74 $\pm$ 0.83}&
    90.08 $\pm$ 0.70&
    82.76 $\pm$ 0.61&
    \textcolor{violet}{78.86 $\pm$ 0.92}&
    35.92 $\pm$ 1.32&
    40.61 $\pm$ 2.97&
    \\

    \multirow{1}[0]{*}
    GloGNN&
    59.63 $\pm$ 0.69&
    36.89 $\pm$ 0.14&
    51.08 $\pm$ 1.23&
    73.39 $\pm$ 1.17&
    65.74 $\pm$ 1.19&
    35.11 $\pm$ 1.24&
    25.90 $\pm$ 3.58&
    \\

    \multirow{1}[0]{*}
    FAGCN&
    65.22 $\pm$ 0.56&
    44.12 $\pm$ 0.30&
    88.17 $\pm$ 0.73&
    77.75 $\pm$ 1.05&
    77.24 $\pm$ 1.26&
    \blue{41.08 $\pm$ 2.27}&
    \blue{41.90 $\pm$ 2.72}&
    \\

    \multirow{1}[0]{*}
    GBK-GNN&
    74.57 $\pm$ 0.47&
    45.98 $\pm$ 0.71&
    90.85 $\pm$ 0.58&
    81.01 $\pm$ 0.67&
    74.47 $\pm$ 0.86&
    35.51 $\pm$ 1.65&
    39.61 $\pm$ 2.60&
    \\

    \multirow{1}[0]{*}
    JacobiCov&
    71.14 $\pm$ 0.42&
    43.55 $\pm$ 0.48&
    89.66 $\pm$ 0.40&
    68.66 $\pm$ 0.65&
    73.88 $\pm$ 1.16&
    29.71 $\pm$ 1.66&
    39.00 $\pm$ 4.20
    \\
    
    \midrule
    \multirow{1}[0]{*}
    DGAT&
    \textcolor{violet}{87.27 $\pm$ 0.64}&
    48.03 $\pm$ 0.58&
    93.27 $\pm$ 0.56&
    \blue{84.74 $\pm$ 0.59}&
    \red{79.55 $\pm$ 0.81}&
    \red{42.09 $\pm$ 2.65}&
    \red{44.16 $\pm$ 4.20}&
    \\

    \multirow{1}[0]{*}
    \textbf{DGAT-sep}&
    \red{89.23 $\pm$ 0.56}&
    50.60 $\pm$ 0.88&
    \red{94.03 $\pm$ 0.45}&
    \red{84.83 $\pm$ 0.40}&
    \blue{78.88 $\pm$ 0.94}&
    \textcolor{violet}{39.69 $\pm$ 2.28}&
    \textcolor{violet}{41.15 $\pm$ 4.66}&
    \\

  \bottomrule
  \end{tabular}
  }
  \caption{Experiment results on heterophily datasets proposed by \cite{platonov2023critical}.
  Values stand for mean and standard deviation of evaluation metrics on the test datasets.
  Here
  \textit{roman-empire}, \textit{amazon-ratings}, \textit{squirrel-filtered} and \textit{chameleon-filtered} use accuracy for evaluation,
  while \textit{minsweeper}, \textit{tolokers} and \textit{questions} use ROC AUC.
  The ``sep" refers to the trick proposed in \cite{zhu2020beyond} which concatenates node's and the mean of neighbours' embedding in each aggregation step, instead of adding them together.
  For fair comparison,
  we use the same experimental setup in \cite{platonov2023critical} for DGAT and DGAT-sep.
  Results for other models are reported by \cite{platonov2023critical}.
  The top three results are highlighted in \red{red}, \blue{blue}, and \textcolor{violet}{violet}, respectively.}
  \label{table:heterophily_new}
    
\end{table*}
\subsection{Synthetic Experiments}
\label{sec:synthetic}
%\subsubsection{Synthetic Datasets}
We follows \cite{Karimi_2018, abu2019mixhop} to generate synthetic graphs characterized by a homophily coefficient $\mu \in \{0.0, 0.1, \ldots, 0.9 \}$, representing the chance that a node forms a connection to another node with the same label. We say the graph is more heterophilic for a smaller $\mu$. 
A detailed explanation about how the synthetic graph are generated is given in  Appendix \ref{appendix:syn-dataset}. We generate 5 synthetic graphs under each homophily coefficient $\mu$. Node features are sampled from overlapping multi-Gaussian distributions. And for each generated graph, nodes are randomly partitioned into train/validation/test sets with a ratio of 60\%/20\%/20\%. Each model is trained under the same hyperparameters setting with the learning rate of $0.01$, the weight decay of $0.001$ and the dropout rate of $0.1$. The number of layers is set to be 2 for each model. And the attention-based model all use 8 heads with 8 hidden states per head.
%while GCN and MLP use 64 hidden states.

%The goal is to investigate the behavior of DGAT in different homophily level, emphasizing both its performance and the optimal choice of $\gamma$. 
%Note that the correlation between the homophily coefficient $\mu$ and $H_{\text{node}}$ is close to linear.
In Table \ref{table:syn_res}, the performance of DGAT are shown along with the fine-tuned hyperparameter $\gamma$ in the bracket, while both the results under the optimal $\gamma^{*}$ and the worst $\gamma^{-}$ are presented.
%\sitao{Did you mention $\gamma^{-}$ before?}.
The DGAT$(\gamma^{*})$ outperforms remarkably better than the original GAT and GATv2 across all heterophily levels.
Even with the worst $\gamma^{-}$, DGAT$(\gamma^{-})$ still achieve comparable or better results than baselines.

We can see that, as the graph becomes more heterophilic with a smaller $\mu$, the optimal $\gamma^{*}$ also decreases. Interestingly, $\gamma^{-}$ has demonstrated a reverse correlation with $\mu$. It aligned with Theorem \ref{thm:gradient-v2} that if the graph has stronger heterophilic property, a large diffusion reduction with a smaller $\gamma$ is preferred. 

This observation provides a general guidance in searching for $\gamma^{*}$. See more detailed discussion about $\gamma$ in Appendix \ref{appendix:choice_of_gamma_synthetic_graphs}.

\subsection{Experiments on Real-world Datasets}
\label{sec:benchmark}
In this subsection, we compare DGAT with 6 baseline models: GAT \cite{velivckovic2017graph}, GAT-sep \footnote{"-sep" means to concatenate the ego feature of node and the aggregated neighborhood information, which is the trick used in \cite{platonov2023critical}, We follow this setting.},
GCN \cite{kipf2016semi},
SAGE \cite{hamilton2017inductive},
Graph Transformer (GT) \cite{shi2021masked} and
GT-sep
and 8 heterophily-specific SOTA models:
H$_2$GCN \cite{zhu2020beyond},
CPGNN \cite{zhu2021graph},
GPR-GNN \cite{chien2020adaptive},
FSGNN \cite{maurya2022simplifying},
GloGNN \cite{li2022finding},
FAGCN \cite{bo2021beyond},
GBK-GNN \cite{du2022gbk} and
JacobiCov \cite{wang2022powerful}
% Tables
% summarizes the empirical results of node classification tasks on real-world datasets.
% \subsubsection{Real-world Datasets}
% \iffalse
% We use datasets
% \textit{Chameleon},
% \textit{Squirrel},
% \textit{Actor},
% \textit{Cornell},
% \textit{Wisconsin},
% \textit{Texas},
% \textit{Cora},
% \textit{Citeseer} and
% \textit{Pubmed}
% \fi
%We compare DGAT against GAT, GATv2 and DGN on 
%\textit{Chameleon},
%\textit{Squirrel},
%\textit{Actor},
%\textit{Cornell},
%\textit{Wisconsin},
%\textit{Texas},
%\textit{Cora},
%\textit{Citeseer} and
%\textit{Pubmed}.
on 7 heterophily recently proposed benchmark datasets \cite{platonov2023critical}:
\textit{roman-empire},
\textit{amazon-ratings},
\textit{minesweeper},
\textit{tolokers},
\textit{questions},
\textit{Chameleon-filtered} and \textit{Squirrel-filtered}.\footnote{The overall statistics of these real-word datasets are given in Appendix \ref{appendix:real-dataset}. We put the comparison results on the datasets from \cite{pei2020geom} in Appendix \ref{app:nine-benchmark}. The reasons that we did show them in the main paper is that, those datasets are proved to have serious drawbacks, including duplicated nodes, small graph size and extreme class imbalance \cite{platonov2023critical}. Thus, the reliability of the comparison results is questionable.}

For the seven new heterophily datasets, we evaluate DGAT under the same experiment setting and fixed splits used \cite{platonov2023critical} with learning rate of $3 \cdot 10^{-5}$, weight decay of 0, dropout rate of 0.2, hidden dimension of 512 and attention head of 8, and number of layers from 1 to 5. Following \cite{platonov2023critical}, DGAT is also trained for 1000 steps with Adam optimizer and select the best step based on the performance on the validation set.

The results are presented in Table \ref{table:heterophily_new}, where GNNs baselines are organized in the beginning block, heterophily-specific GNNs are put in the second block. 

Basically,
DGAT achieve performance improvement compared with its base-model GAT (and DGAT-sep outperforms GAT-sep).
And DGAT consistently surpasses SOTA GNNs that are specifically designed to deal with heterophily problems.
While incorporating the ``sep"-trick further enhances DGAT's performance on \textit{roman-empire}, \textit{minesweeper} and \textit{tolokers}.
Since the ``sep"-trick allows the aggregation step to put negative weights on the propagated message,
which enables node-wise diversification that is proved to be useful on heterophily data \cite{luan2022revisiting}. 
In addition,
the outstanding of DGAT from GT-based methods
indicates that finding the useful graph for message propagation with sparse attention is more effective than considering the message passing between all pairs of nodes.

\section{Conclusion}
In this work, we propose a novel parameterized normalized Laplacians, encompassing both the normalized Laplacian and the combinatorial Laplacian.
With our proposed Laplacian, 
we design a Directional Graph Attention Network (DGAT) with controllable global directional flow
that substantially enhances the expressiveness of the original Graph Attention Network (GAT) on a general graph.
% Building upon the global directional flow, we proposed the Directional Graph Attention Network (DGAT), which is an attention-based graph neural network enriched with embedded global directional information.
The two mechanisms proposed in this study have contributed to the improvement:
% More specially, we proposed two mechanisms on top of the GAT: 
the topology-guided neighbour rewiring and the global directional attention mechanism.
The future research based on our work will focus on developing a quantitative guidance for selecting the parameter $\gamma$ in the proposed parameterized normalized Laplacians.

\iffalse
\section{Impact Statements}
This paper presents work whose goal is to advance the field of Machine Learning. There are many potential societal consequences of our work, none which we feel must be specifically highlighted here.
\fi

\clearpage
\bibliographystyle{icml2024}
\bibliography{example_paper}

\clearpage
\appendix
\section{Homophily Metrics}
\label{app:metric}
In addition to the node homophily metric introduced in \eqref{eq:node-homo},
here we review some commonly used metrics to measure homophily.
We denote $\z \in \Rbb^N$ as labels of nodes,
and $\mathbf{Z} \in \Rbb^{N\times C}$ as the one-hot encoding of labels,
where $C$ is the number of classes.
Edge homophily $H_{\mathrm{edge}}$ is defined as follows:
\begin{align}
H_{\mathrm{edge}} = \frac{\{ e_{ij}| e_{ij} \in \mathcal{E}, \z_i = \z_j \}}{|\mathcal{E}|}.
\end{align}
Adjusted edge homophily $H_{\mathrm{edge}}^{*}$
considers classes imbalance and is defined as \cite{platonov2023critical} :
\begin{align}
H_{\mathrm{edge}}^{*} = \frac{H_{\mathrm{edge}} - \sum_c p^2(c)}{1- \sum_c p^2(c)}.
\end{align}
Here $p(c) = \sum_{i: \z_i=c} \D_{ii} / (2 |\mathcal{E}|)$, $c=1:C$, defines the degree-weighted distribution of class labels.
The class homophily is also proposed to take class imbalance into account \cite{lim2021new}:
\begin{align}
&H_{\mathrm{class}} = \frac{1}{C-1}\sum_{c}\left[h_c - \frac{|\{v_i|\z_i = c\}|}{N} \right]_{+}\\   
&h_c = \frac{\sum_{v_i: \z_i = c} |\{e_{ij}|  e_{ij} \in \mathcal{E}, \z_i = \z_j \}|}{\sum_{v_i: \z_i = c} \D_{ii}}.
\end{align}
Label informativeness,
which indicates the amount of information a neighbor's label provides about node's label,
is defined as follows \cite{platonov2023critical}:
\begin{align}
LI = 2 - \frac{\sum_{c_1, c_2} p(c_1, c_2) \log p(c_1, c_2)}{\sum_c p(c) \log p(c)},  
\end{align}
where $p(c_1, c_2) = |\{e_{ij}|e_{ij} \in \mathcal{E}, \z_i = c_1, \z_j = c_2 \}|/(2|\mathcal{E}|)$.
The aggregation homophily $H_{\mathrm{agg}}^{\mathrm{M}}(\mathcal{G})$ measures the proportion of nodes that assign greater average weights to intra-class nodes than inter-class nodes.
It is defined as follows \cite{luan2022revisiting}:
\begin{align}
H_{\mathrm{agg}}^{\mathrm{M}}(\mathcal{G}) = \frac{1}{|\mathcal{V}|} \big| \{ &v_i| 
\mathrm{Mean}_{j} (\{S(\hat{\mathbf{A}}, \Z)_{ij}|\z_i=\z_j \} ) \\
& \geq \mathrm{Mean}_{j}(\{S(\hat{\mathbf{A}}, \Z)_{ij}|\z_i\neq \z_j \} )\} \big| \nonumber, 
\end{align}
where $S(\hat{\mathbf{A}}, \Z) = \hat{\mathbf{A}}\Z (\hat{\mathbf{A}}\Z)^\top$ defines the post-aggregation node similarity,
with $\hat{\mathbf{A}} = \mathbf{A} + \mathbf{I}$,
and $\mathrm{Mean}_{j}(\{\cdot\})$ takes the average over node $v_j$ of a given multiset of values.
Under the homophily metrics mentioned above,
a smaller value  
indicates a higher degree of heterophily.
While $H_{\mathrm{edge}}^{*}$ can assume negative values,
the other metrics fall within the range $[0, 1]$.

% We review the theoretical analysis in \cite{beaini2021directional}, which justifies the choice of using the eigenvector $\phib^{(1)}$ corresponding to the smallest non-trivial eigenvalue $\lambda^{(1)}$
% of a graph Laplacian to define the global directional flow in a graph.
% This analysis also serves as an important theoretical foundation to our work.
 
% The theorem in \cite{beaini2021directional} states that by following the gradient of the eigenvectors, the diffusion  distance between a pair of nodes on a graph could be reduced effectively.

%\section{Additional definition and theorem}

%When $\alpha=1$, the following result indicates that $\P^{(1, \gamma)}$ is a valid random walk matrix.

\section{Proof of Theorem}
\subsection{Proof of theorem \ref{thm:parameterized-matrix}}
\label{appendix:proof_parameterized_matrix}
\begin{proof}
 By Definition \ref{def:aug-lap}, we have
\begin{align*}
& \resizebox{1\hsize}{!}{$\P^{(\alpha, \gamma)} = \I - \mathbf{L}^{(\alpha,\gamma)}
    = \I - \gamma [\gamma \D + (1-\gamma)\I]^{-\alpha} \mathbf{L}[\gamma\D + (1-\gamma)\I]^{\alpha-1} $} \\
& \resizebox{1\hsize}{!}{$=[\gamma \D + (1-\gamma)\I]^{-\alpha}
  [\gamma \D + (1-\gamma)\I - \gamma \mathbf{L}]
 [\gamma \D + (1-\gamma)\I]^{\alpha-1} $} \\
& \resizebox{1\hsize}{!}{$ =[\gamma \D + (1-\gamma)\I]^{-\alpha}
  [\gamma \A + (1-\gamma)\I]
 [\gamma \D + (1-\gamma)\I]^{\alpha-1}  $.}
\end{align*}
It is easy to see that all elements in $\P^{(\alpha,\gamma)}$ are non-negative. 
%When $\alpha=1$, 
% $$
%     \P^{(1,\gamma)}
%     = [\gamma\D + (1-\gamma)\I]^{-1}
%     [\gamma\A+(1-\gamma)\I ]
% $$
Since $\A\1=\D\1$, we have
$$
\P^{(1,\gamma)} \1 = (\gamma\D + (1-\gamma)\I)^{-1}
        \left(\gamma\A+(1-\gamma)\I\right) \1 = \1,
$$
%then $\P^{(1, \gamma)}$ is a valid random walk matrix.
completing the proof.
\end{proof}

\subsection{Proof of theorem \ref{thm:eigenplaplaican}}
\label{appendix:proof_eigenplaplaican}
\begin{proof}
For any nonzero $\x\in \Rbb^N$, 
write $\y:=[\gamma \D + (1-\gamma)\I]^{-1/2}\x$.
Then we have
\begin{align*}
& \resizebox{1\hsize}{!}{$\frac{\x^\top \mathbf{L}^{(1/2, \gamma)}\x}{\x^\top\x} = \frac{\x^\top\gamma[\gamma \D + (1-\gamma)\I]^{-1/2} \mathbf{L}
[\gamma \D + (1-\gamma)\I]^{-1/2}\x}{\x^\top\x} $} \\
& \resizebox{1\hsize}{!}{$= \frac{\gamma\y^\top \mathbf{L} \y}{\y^\top [\gamma \D + (1-\gamma)\I] \y} =\frac{\frac{\gamma}{2} \sum_{ij} a_{ij}(y_i-y_j)^2}
{\gamma \sum_{ij} a_{ij}y_i^2 + (1 - \gamma) \sum_{i}y_i^2}$.}
\end{align*}
By the Rayleigh quotient theorem,  
\be \label{eq:rayleighr}
\lambda^{(0)}(\gamma)
=\min_{\y \neq \0}  \frac{ \frac{\gamma}{2}\sum_{ij} a_{ij}(y_i-y_j)^2}
{\gamma\sum_{ij} a_{ij}y_i^2+ (1 - \gamma) \sum_{i}y_i^2} = 0,
\ee
where the minimum is reached when $\y$ is a multiple of $\1$
and
\begin{align*}
\lambda^{(N-1)}(\gamma)
& = \max_{\y \neq \0} \frac{ \frac{\gamma}{2}\sum_{ij} a_{ij}(y_i-y_j)^2}
{\gamma\sum_{ij} a_{ij}y_i^2+ (1 - \gamma)\sum_{i}y_i^2}  \\
&  \leq  \max_{\y \neq \0} \frac{ \sum_{ij} a_{ij}  (y_i^2 +y_j^2)}
 {\sum_{ij} a_{ij}y_i^2}  
 \leq 2,
\end{align*}
leading to Eq.~\eqref{eq:glevrange}.
The proof of showing $\lambda^{(1)}(\gamma)\neq 0$ if and only if ${\cal G}$ is connected is similar to \cite{kim2016algebraic}, thus we omit the details here.

By the Courant-Fischer min-max theorem,  
for $\gamma\neq 0$,
\begin{align*}
    \lambda^{(i)}(\gamma)
& = \min_{\{S:\text{dim}(S)=i+1\}} \max_{\{\x:\0\neq \x\in S\}} 
\frac{\x^\top \mathbf{L}^{(1/2, \gamma)}\x}{\x^\top\x} \\
& = \min_{\{S:\text{dim}(S)=i+1\}}\max_{\{\y:\0\neq \y\in S\}} 
\frac{\y^\top\mathbf{L}\y}{\y^\top[\D-\I+ (1/\gamma)\I]\y}.
\end{align*}
It is obvious that the Rayleigh quotient
$\frac{\y^\top\mathbf{L}\y}{\y^\top [\D-\I+ (1/\gamma)\I] \y}$
is strictly increasing with respect to $\gamma\in (0,1]$ if $\mathbf{L}\y\neq 0$,
\ie{} $\y$ not a multiple of $\1$. 
Note that $\lambda^{(0)}(\gamma)=0$ and it is reached when 
$\mathbf{L}\y=\0$, or  equivalently $\y$ is a multiple of $\1$.
Thus, $\lambda^{(i)}(\gamma)$ is strictly increasing with respect to $\gamma$
for $i=1:N-1$.

From the eigendecomposition of the symmetric $\mathbf{L}^{(1/2,\gamma)}$ in ~\eqref{eq:gled},
we can find the eigendecomposition of $\mathbf{L}^{(\alpha,\gamma)}$
as follows:
\begin{align*}
&\resizebox{1\hsize}{!}{$ \mathbf{L}^{(\alpha,\gamma)} = [\gamma\D+(1-\gamma)\I]^{1/2-\alpha}\mathbf{L}^{(1/2, \gamma)} [\gamma\D+(1-\gamma)\I]^{\alpha -1/2} $} \\
& \resizebox{1\hsize}{!}{$ = [\gamma\D+(1-\gamma)\I]^{1/2-\alpha}(\U \bLambda^{(\gamma)} \U^\top)
[\gamma\D+(1-\gamma)\I]^{\alpha -1/2} $}\\
& \resizebox{1\hsize}{!}{$ = \left([\gamma\D+(1-\gamma)\I]^{1/2-\alpha}\U\right) \bLambda^{(\gamma)}  
\left([\gamma\D+(1-\gamma)\I]^{1/2-\alpha}\U\right)^{-1} $}
\end{align*}
Thus,  $\lambda^{(i)} (\gamma)$ is also 
an eigenvalue of $\mathbf{L}^{(\alpha,\gamma)}$ for $i=0:N-1$,
and the $i$-th column of $[\gamma\D+(1-\gamma)\I]^{1/2-\alpha}\U$ is
a corresponding eigenvector.
\end{proof}

\subsection{Proof of Theorem \ref{thm:gradient-v2}}
\label{appendix:gradient-v2}
\begin{proof}
The proof is similar to the proof of \cite{beaini2021directional}.
%Let $p_k(v_i, v_j) = \left(\P^{(1, \gamma)} \right)^{k}(i, j)$,
%then $q_k(v_i, v_j) = \sum^{\infty}_{n=0} \frac{e^{-t}t^{k}}{k!}p_k(v_i, v_j)$,
%which is the transition probability from node $v_i$ to $v_j$ as defined in Eq.~\eqref{eq:transit-prob}.
%The parameterized diffusion distance can be written as
% \be
% \label{eq:pf-dt}
%    d_t(v_i, v_j) \coloneqq 
%     \left( \sum_{v_m \in \mathcal{V}} \left(q_t(v_i, v_m) - q_t(v_j, v_m)
%     \right)^2 \right)^\frac{1}{2}.
% \ee
By \cite{Coifman2006Diffusion}, 
the diffusion distance at time $t$ between node $v_i$ and $v_j$ can be expressed as:
%we can rewrite Eq.~\eqref{eq:pf-dt} as:
    \be
    \label{eq:pf-dt2}
       d_t(v_i, v_j) = \left( \sum_{k=1}^{n-1} e^{-2t \lambda^{(k)}(\gamma)}(\phi_i^{(k)}(\gamma) - \phi_j^{(k)}(\gamma))^2 \right)^\frac{1}{2},
    \ee
    where 
    $
    \lambda^{(1)}(\gamma) \leq \lambda^{(2)}(\gamma) \leq \dots \leq \lambda^{(n-1)}(\gamma)
    $
    are eigenvalues of $\mathbf{L}^{(1, \gamma)}$, and $\{\phib^{(1)}(\gamma), \phib^{(2)}(\gamma), \dots, \phib^{(n-1)}(\gamma)\}$ are the corresponding eigenvectors. We omit the zero $\lambda^{(0)}(\gamma)$.
    The inequality $d_t(v_m, v_j) < d_t(v_i, v_j)$ is then equivalent as 
    \be
        \begin{aligned}
        \left( \sum_{k=1}^{n-1} e^{-2t \lambda^{(k)}(\gamma)}(\phi_m^{(k)}(\gamma) - \phi_j^{(k)}(\gamma))^2 \right)^\frac{1}{2}&\\
        < \left( \sum_{k=1}^{n-1} e^{-2t \lambda^{(k)}(\gamma)}(\phi_i^{(k)}(\gamma) - \phi_j^{(k)}(\gamma))^2
        \right)^\frac{1}{2}.
        \end{aligned}
    \ee
    We can take out $\lambda^{(1)}(\gamma)$ and $\phib^{(1)}(\gamma)$ and rearrange the above inequality as:
    \be
        \begin{aligned}
            \label{eq:pf-ineq}
            &\resizebox{1\hsize}{!}{$\sum_{k=2}^{n-1} e^{-2t \lambda^{(k)}(\gamma)}
            \left(
            (\phi_m^{(k)}(\gamma) - \phi_j^{(k)}(\gamma))^2 - 
            (\phi_i^{(k)}(\gamma) - \phi_j^{(k)}(\gamma))^2 
            \right) $}\\
            &\resizebox{1\hsize}{!}{$ < e^{-2t \lambda^{(1)}(\gamma)}\left(
            (\phi_i^{(1)}(\gamma) - \phi_j^{(1)}(\gamma))^2 - 
            (\phi_m^{(1)}(\gamma) - \phi_j^{(1)}(\gamma))^2 
            \right).$}
        \end{aligned}
    \ee
    The left-hand side of Eq.~\eqref{eq:pf-ineq} 
    %is bounded above by:
    has an upper bound:
    \be
    \begin{aligned}
         \resizebox{1\hsize}{!}{$\sum_{k=2}^{n-1} e^{-2t \lambda^{(k)}(\gamma)}
        \left|
        (\phi_m^{(k)}(\gamma) - \phi_j^{(k)}(\gamma))^2 - 
        (\phi_i^{(k)}(\gamma) - \phi_j^{(k)}(\gamma))^2 
        \right| $}\\
        \resizebox{1\hsize}{!}{$\leq
        e^{-2t \lambda^{(2)}(\gamma)}
        \sum_{k=2}^{n-1} 
        \left|
        (\phi_m^{(k)}(\gamma) - \phi_j^{(k)}(\gamma))^2 - 
        (\phi_i^{(k)}(\gamma) - \phi_j^{(k)}(\gamma))^2 
        \right|.$}
    \end{aligned}
    \ee
    Then Eq.~\eqref{eq:pf-ineq} holds if:
    \be
        \begin{aligned}
            \label{eq:pf-ineq2}
             \resizebox{1\hsize}{!}{$e^{-2t \lambda^{(2)}(\gamma)}
            \sum_{k=2}^{n-1} 
            \left|
            (\phi_m^{(k)}(\gamma) - \phi_j^{(k)}(\gamma))^2 - 
            (\phi_i^{(k)}(\gamma) - \phi_j^{(k)}(\gamma))^2 
            \right|$}\\
            \resizebox{1\hsize}{!}{$\leq
            e^{-2t \lambda^{(1)}(\gamma)}\left(
            (\phi_i^{(1)}(\gamma) - \phi_j^{(1)}(\gamma))^2 - 
            (\phi_m^{(1)}(\gamma) - \phi_j^{(1)}(\gamma))^2 
            \right),$}
        \end{aligned}
    \ee
    which is equivalent to 
    
 \be
    \begin{aligned}
        \label{eq:pf-ineq3}
        &\log\left(
        \frac
        {(\phi_i^{(1)}(\gamma) - \phi_j^{(1)}(\gamma))^2 - 
        (\phi_m^{(1)}(\gamma) - \phi_j^{(1)}(\gamma))^2}
        {\sum_{k=2}^{n-1} 
        \left|
        (\phi_m^{(k)}(\gamma) - \phi_j^{(k)}(\gamma))^2 - 
        (\phi_i^{(k)}(\gamma) - \phi_j^{(k)}(\gamma))^2 
        \right|}
        \right)\\
        & \times \frac{1}{2(\lambda^{(1)}(\gamma) - \lambda^{(2)}(\gamma))} < t.
    \end{aligned}
    \ee      
    Let the constant $C$ be the left-hand side of Eq.~\eqref{eq:pf-ineq3}, 
    then if we take $t \geq \left \lfloor C \right \rfloor + 1$, 
    we have $d_t(v_m, v_j) < d_t(v_i, v_j)$.    
    Note that $C$ exits if
    \be
        \frac{(\phi_i^{(1)}(\gamma) - \phi_j^{(1)}(\gamma))^2 - 
        (\phi_m^{(1)}(\gamma) - \phi_j^{(1)}(\gamma))^2}
        {\sum_{k=2}^{n-1} 
        \left|
        (\phi_m^{(k)}(\gamma) - \phi_j^{(k)}(\gamma))^2 - 
        (\phi_i^{(k)}(\gamma) - \phi_j^{(k)}(\gamma))^2 
        \right|} > 0,
    \ee
    which is satisfied since we assume
    $| \phi_i^{(1)}(\gamma) - \phi_j^{(1)}(\gamma) | > | \phi_m^{(1)}(\gamma) - \phi_j^{(1)}(\gamma)|$.
    % and is based on the assumption that the chosen neighbor $v_m$ of $v_i$ always satisfies the condition that $\phi_i^{(1)}(\gamma) < \phi_m^{(1)}(\gamma) < \phi_j^{(1)}(\gamma)$.
    The original theorem \cite{beaini2021directional} is only based on $\phib^{(1)}$ and does not assume that $v_m$ must satisfy $| \phi_i^{(1)} - \phi_j^{(1)} | > | \phi_m^{(1)} - \phi_j^{(1)}|$,
    which is necessary for the existence of $C$.
    In addition, the original theorem \cite{beaini2021directional} assumes that $v_m$ is obtained by taking a gradient step from $v_i$, \ie{} $\phi_m - \phi_i = \max_{j:v_j \in \calN(v_i)} (\phi_j - \phi_i)$,
    while this property is not needed for the proof.
    Therefore,
    Theorem \ref{thm:gradient-v2} both extends and overcomes the shortcomings of \cite{beaini2021directional}.

\end{proof}

\section{Datasets}
\subsection{Real-world Datasets}
\label{appendix:real-dataset}
The overall statistics of the real-world datasets are presented in Table \ref{table:chars} and
Table \ref{table:heter_metric} provides their heterophily levels calculated using various homophily metrics.
% In the following sub-section, we give a thorough overview of each of them.
\begin{table}[htbp]
  \centering  
  \resizebox{1\hsize}{!}{
  \begin{tabular}{lcccccc}
  \toprule
  &
  \#Nodes&
  \#Edges&
  \#Features&
  \#Classes&
  Metric
  \\

  \midrule
  chameleon&
  2,277&
  31,371&
  2,325&
  5&
  ACC
  \\

  squirrel&
  5,201&
  198,353&
  2,089&
  5&
  ACC
  \\

  actor&
  7,600&
  26,659&
  932&
  5&
  ACC
  \\

  cornell&
  183&
  277&
  1,703&
  5&
  ACC
  \\

  wisconsin&
  251&
  450&
  1,703&
  5&
  ACC
  \\

  texas&
  183&
  279&
  1,703&
  5&
  ACC
  \\

  cora&
  2,708&
  5,278&
  1,433&
  7&
  ACC
  \\

  citeseer&
  3,327&
  4,552&
  3,703&
  6&
  ACC
  \\

  pubmed&
  19,717&
  44,324&
  500&
  3&
  ACC
  \\

  roman-empire&
  22,662&
  32,927&
  300&
  18&
  ACC
  \\

  amazon-ratings&
  24,492&
  93,050&
  300&
  5&
  ACC
  \\

  minesweeper&
  10,000&
  39,402&
  7&
  2&
  ROC AUC
  \\

  tolokers&
  11,758&
  519,000&
  10&
  2&
  ROC AUC
  \\

  questions&
  48,921&
  153,540&
  301&
  2&
  ROC AUC
  \\

  squirrel-filtered&
  2,223&
  46,998&
  2,089&
  5&
  ACC
  \\

  chameleon-filtered&
  890&
  8,854&
  2,325&
  5&
  ACC
  \\

  \bottomrule
  \end{tabular}
  }
    \caption{
    Statistics of the benchmark dataset. Following pre-processing, the graph has been transformed into an undirected and simple form, without self-loops or multiple edges.}
\end{table}
\label{table:chars}
\begin{table}[htbp]
  \centering  
  \resizebox{1\hsize}{!}{
  \begin{tabular}{lcccccc}

  \toprule
  &
  $H_{\mathrm{node}}$&
  $H_{\mathrm{edge}}$&
  $H_{\mathrm{class}}$&
  $H_{\mathrm{agg}}^{\mathrm{M}}(\mathcal{G})$&
  $H_{\mathrm{edge}}^{*}$&
  LI
  \\

  \midrule
  chameleon&
  0.25 & % h node
  0.23 & % h edge
  0.04 & % h class
  0.36 & % h agg
  0.03 & % adjusted h edge
  0.05  % LI

  \\
  
  squirrel&
  0.22& % h node
  0.22& % h edge
  0.03& % h class
  0.00& % h agg
  0.01& % adjusted h edge
  0.00  % LI

  \\

  actor&
  0.22& % h node
  0.22& % h edge
  0.01& % h class
  0.62& % h agg
  0.00& % adjusted h edge
  0.00  % LI

  \\

  cornell&
  0.30& % h node
  0.30& % h edge
  0.02& % h class
  0.02& % h agg
  -0.08& % adjusted h edge
  0.02 % LI

  \\

  wisconsin&
  0.16& % h node
  0.18& % h edge
  0.05& % h class
  0.00& % h agg
  -0.17& % adjusted h edge
  0.13  % LI

  \\

  texas&
  0.06& % h node
  0.06& % h edge
  0.00& % h class
  0.00& % h agg
  -0.29& % adjusted h edge
  0.19  % LI
  \\

  cora&
  0.83& % h node
  0.81& % h edge
  0.77& % h class
  0.99& % h agg
  0.77& % adjusted h edge
  0.59 % LI

  \\

  citeseer&
  0.71& % h node
  0.74& % h edge
  0.63& % h class
  0.97& % h agg
  0.67& % adjusted h edge
  0.45 % LI

  \\

  pubmed&
  0.79& % h node
  0.80& % h edge
  0.66& % h class
  0.94& % h agg
  0.69& % adjusted h edge
  0.41  % LI

  \\
  roman-empire&
  0.05& % h node
  0.05& % h edge
  0.02& % h class
  1.00& % h agg
  -0.05& % adjusted h edge
  0.11  % LI
  \\

  amazon-ratings&
  0.38& % h node
  0.38& % h edge
  0.13& % h class
  0.60& % h agg
  0.14& % adjusted h edge
  0.04  % LI
  \\

  minesweeper&
  0.68& % h node
  0.68& % h edge
  0.01& % h class
  0.61& % h agg
  0.01& % adjusted h edge
  0.00  % LI
  \\

  tolokers&
  0.63& % h node
  0.59& % h edge
  0.18& % h class
  0.00& % h agg
  0.09& % adjusted h edge
  0.01  % LI 
  \\

  questions&
  0.90&
  0.84&
  0.08&
  0.00&
  0.02&
  0.00
  \\

  squirrel-filtered&
  0.19& % h node
  0.21& % h edge
  0.04& % h class
  0.00& % h agg
  0.01& % adjusted h edge
  0.00  % LI 
  \\

  chameleon-filtered&
  0.24& % h node
  0.24& % h edge
  0.04& % h class
  0.25& % h agg
  0.03& % adjusted h edge
  0.01  % LI 
  \\
  
  \bottomrule
  \end{tabular}
  }
  \caption{Heterophily levels of benchmark datasets. The $H_{\mathrm{agg}}^{\mathrm{M}}(\mathcal{G})$ stands for the aggregation homophily, calculated using $\hat{\mathbf{A}} = \mathbf{A} + \mathbf{I}$. The $H_{\mathrm{edge}}^{*}$ stands for the adjusted edge homophily, and the LI stands for the label informativeness. 
  The definitions of these metrics can be found in \eqref{eq:node-homo} and in Appendix \ref{app:metric}.}
\end{table}
\label{table:heter_metric}

\subsection{Synthetic Datasets}
\label{appendix:syn-dataset}
In addition to the real-world datasets, we also tested the DGAT model on synthetic graphs generated with different homophily levels ranging from 0 to 1 using the method proposed in \cite{abu2019mixhop}.
Here we give a review of the generation process.

More specifically, 
when generating the output graph $\mathcal{G}$ with a desired total number of nodes $N$, a total of $C$ classes,
and a homophily coefficient $\mu$,
the process begins by dividing the $N$ nodes into $C$ equal-sized classes.
Then the synthetic graph $\mathcal{G}$ (initially empty) is updated iteratively.
At each step,
a new node $v_i$ is added,
and its class $z_i$ is randomly assigned from the set $\{1, \ldots, C\}$.
Whenever a new node $v_i$ is added to the graph, we 
establish a connection between it and an existing node $v_j$ in $\mathcal{G}$ based on the probability $p_{ij}$ determined by the following rules:
\be
\label{eq:puv}
p_{ij} = \begin{cases}
d_j \times \mu, & \mbox{if $z_i = z_j$} \\
d_j \times (1-\mu) \times w_{d(z_i, z_j)}, & \mbox{otherwise}
\end{cases}.
\ee
where $z_i$ and $z_j$ are class labels of node $i$ and $j$ respectively, and $w_{d(z_i, z_j)}$ denotes the ``cost'' of connecting nodes from two distinct classes with a class distance of $d(z_i, z_j)$.
For a larger $\mu$, the chance of connecting with a node with the same label increases.
The distance between two classes simply implies the shortest distance between the two classes on a circle where classes are numbered from 1 to $C$. 
For instance, if $C = 6$, $z_i = 1$ and $z_j = 5$, then the distance between $z_i$ and $z_j$ is $2$. 
The weight exponentially decreases as the distance increases and is normalized such that $\sum_d w_d = 1$.
In addition, the probability $p_{ij}$ defined in Eq.~\eqref{eq:puv} is also normalized over the exiting nodes:
$$
    \bar{p}_{ij} = \frac{p_{ij}}{\sum_{k: v_k \in \mathcal{N}(v_i)} p_{ik}}
$$
Lastly, the features of each node in the output graph are sampled from overlapping 2D Gaussian distributions.
Each class has its own distribution defined separately.

\begin{table*}[htbp]
  %\left 
  \begin{tabular}{l|c|ccccc|c}
  \toprule
  $\mathcal{R}(\mathcal{G})$&
  &
  $\checkmark$&
  $\checkmark$&
  &
  &
  $\checkmark$&
  $\checkmark$

  \\

  $\text{ATT}_\mathbf{e}$&
  &
  &
  &
  $\checkmark$&
  $\checkmark$&
  $\checkmark$&
  $\checkmark$
  
  \\

  $\mathbf{L}^{(\alpha, \gamma)}$&
  &
  & 
  $\checkmark$&
  &
  $\checkmark$&
  &
  $\checkmark$
  
  \\

  \midrule
  chameleon&
  42.93&
  51.71 $\pm$ 1.23&
  51.91 $\pm$ 1.59&
  46.69 $\pm$ 1.98&
  46.95 $\pm$ 2.50&
  52.39 $\pm$ 2.08&
  \textbf{52.47 $\pm$ 1.44}
  
  \\

  squirrel&
  30.03&
  32.91 $\pm$ 2.26&
  34.70 $\pm$ 1.02&
  34.67 $\pm$ 0.10&
  34.71 $\pm$ 1.02&
  33.93 $\pm$ 1.86&
  \textbf{34.82 $\pm$ 1.60}
  
  \\

  actor&
  28.45&
  35.24 $\pm$ 1.07&
  35.51 $\pm$ 0.74&
  29.63 $\pm$ 0.52&
  29.87 $\pm$ 0.60&
  35.64 $\pm$ 0.10&
  \textbf{35.68 $\pm$ 1.20}
  
  \\

  cornell&
  54.32&
  82.16 $\pm$ 5.94&
  84.32 $\pm$ 4.15&
  60.27 $\pm$ 4.02&
  60.54 $\pm$ 0.40&
  84.86 $\pm$ 5.12&
  \textbf{85.14 $\pm$ 5.30}
  
  \\

  wisconsin&
  49.41&
  83.92 $\pm$ 3.59&
  84.12 $\pm$ 3.09&
  54.51 $\pm$ 7.68&
  54.71 $\pm$ 5.68&
  84.11 $\pm$ 2.69&
  \textbf{84.71 $\pm$ 3.59}
  
  \\

  texas&
  58.38&
  78.38 $\pm$ 5.13&
  78.65 $\pm$ 4.43&
  60.81 $\pm$ 5.16&
  60.81 $\pm$ 4.87&
  78.92 $\pm$ 5.57&
  \textbf{79.46 $\pm$ 3.67}
  
  \\

  cora&
  86.37&
  87.85 $\pm$ 1.02&
  87.83 $\pm$ 1.20&
  87.95 $\pm$ 1.01&
  87.28 $\pm$ 1.53&
  87.93 $\pm$ 0.99&
  \textbf{88.05 $\pm$ 1.09}
  
  \\

  citeseer&
  74.32&
  74.50 $\pm$ 1.99&
  76.18 $\pm$ 1.35&
  76.31 $\pm$ 1.38&
  \textbf{76.41 $\pm$ 1.62}&
  73.91 $\pm$ 1.62&
  \textbf{76.41 $\pm$ 1.45}
  
  \\

  pubmed&
  87.62&
  87.78 $\pm$ 0.44&
  87.93 $\pm$ 0.36&
  87.90 $\pm$ 0.40&
  87.88 $\pm$ 0.46&
  87.82 $\pm$ 0.30&
  \textbf{87.94 $\pm$ 0.48}
  
  \\
  \bottomrule
  \end{tabular}
\caption{Ablation study on 9 real-world datasets \cite{pei2020geom}. Cell with \checkmark means the component is applied to the DGAT model.
The $\mathbf{L}^{(\alpha, \gamma)}$ means that the parameterized normalized Laplacian is used with optimal $\alpha$ and $\gamma$,
while unchecking $\mathbf{L}^{(\alpha, \gamma)}$ indicates that the default $\mathbf{L}^{(1, 1)} = \mathbf{L}_{\rw}$ is used.
The $\text{ATT}_\mathbf{e}$ is the global directional attention mechanism.
And $\mathcal{R}(\mathcal{G})$ is the graph re-wiring strategy.
DGAT without all component is just GAT,
whose results are reported from \cite{pei2020geom}.
The best test results are highlighted in \textbf{bold} format.}
\label{table:ablation_study_vertical}
\end{table*}
\begin{figure*}[htbp!]
    \centering
     {  
     {
     \subfloat[PBE]{
     \captionsetup{justification = centering}
     \includegraphics[width=0.5\textwidth]{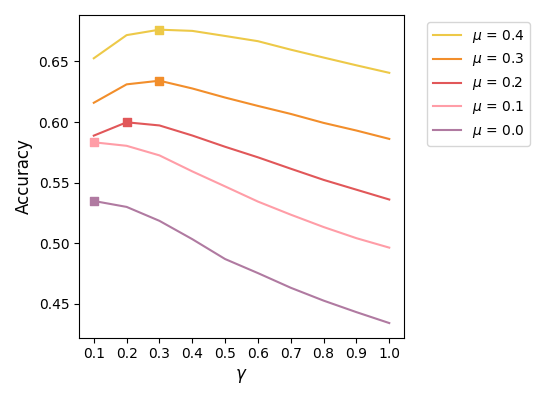}
     }
     \subfloat[$D_\text{NGJ}$]{
     \captionsetup{justification = centering}
     \includegraphics[width=0.5\textwidth]{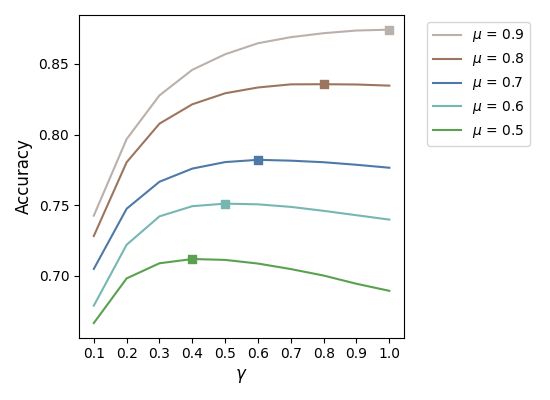}
     } 
     }
     }
       \caption{
The performance of a one layer GCN with aggregation weights defined by $\P^{(1,\gamma)}$ on synthetic graphs. Results are shown separately for heterophily ( $\mu < 0.5$) on the left panel and homophily ($\mu \geq 0.5$) on the right,
    with the solid line as the mean accuracy over 5 splits. The optimal $\gamma$ for each $\mu$ is highlighted using square.
  }
     \label{fig:syn_gcn}
\end{figure*}

\subsection{On the choice of $\gamma$ for synthetic graphs}
\label{appendix:choice_of_gamma_synthetic_graphs}
We use the performance of a 1-layer GCN on the synthetic graphs \footnote{The training use the same setting as Section \ref{sec:synthetic},
where the learning rate of $0.01$, the weight decay of $0.001$, the dropout rate of $0.1$, and the hidden dimension of 64 are used.}
to further illustrate how the hyper-parameter $\gamma$ in $\mathbf{L}^{(1,\gamma)}$ influences the effectiveness of graph message passing under different heterophily levels.
As defined in \eqref{def:adj},
the parameterized normalized adjacent matrix for $\mathbf{L}^{(1,\gamma)}$ is
$$\P^{(1,\gamma)} = (\gamma\D + (1-\gamma)\I)^{-1} \left(\gamma\A+(1-\gamma)\I\right).$$
Then the output of a 1-layer GCN with aggregation weights defined by $\P^{(1,\gamma)}$ can be expressed as 
$$\mathrm{softmax}\big( \mathrm{ReLU} (\P^{(1,\gamma)} \X \W_0) \W_1 \big).$$
As the Figure \ref{fig:syn_gcn} shows,
generally,
the performance of
GCN increases with $\mu$.
The optimal $\gamma$ increases as the graph becomes less homophilic,
which aligns with results in Section \ref{sec:synthetic} and Theorem \ref{thm:gradient-v2}.
Furthermore, in the case of heterophily ($\mu < 0.5$), a larger $\gamma$ tends to have a detrimental effect on performance.
With the increase of $\mu$,
the influence on performance brought by a larger $\gamma$ gradually changes from negative to positive. 

\section{Trainning}
\label{app:training}
In training and evaluating a model using a node classification benchmark dataset with $C$ distinct classes, 
each node $v_i \in \mathcal{V}$ has a label $z_i$ associated with it.
We denote $\mathbf{Z} \in \Rbb^{N \times C}$ as the one-hot encoding of labels.
Moreover, nodes are divided into three sets: the training set $\mathcal{V}_{\mathrm{train}}$, the validation set $\mathcal{V}_{\mathrm{val}}$ and the test set $\mathcal{V}_{\mathrm{test}}$.
In the training phase, 
the model uses features of all nodes under transductive learning.
The model only has access to labels of nodes in $\mathcal{V}_{\mathrm{train}}$ and in $\mathcal{V}_{\mathrm{val}}$ (for hyperparameter tuning),
while labels of nodes in $\mathcal{V}_{\mathrm{test}} = \mathcal{V} \setminus (\mathcal{V}_{\mathrm{train}} \cup \mathcal{V}_{\mathrm{val}})$ remain unknown to the model.

The cost function used in node classification tasks is the standard categorical cross-entropy loss \cite{hamilton2020graph}, which is commonly used for multi-class classification tasks:
\be 
    \mathcal{L} = 
     - \frac{1}{|\mathcal{V}_{\mathrm{train}}|}\mathrm{trace}(\Z^\top \log \Y),
    \label{eq:loss}
\ee
%where $\z_i \in \Rbb^T$ is the one-hot encoding for the ground-truth class label of node $v_i$,
% $
% \Tilde{\h}^{(2)}_i
% = (\softmax(\mathbf{H}^{(2)}_{i, :}))^\top,
% $
where $\Y$ is the output from the model after $\mathrm{softmax}$
and $\log(\cdot)$ is applied element-wise.
% In our experiment, we adopt the batch gradient descent method using the full training dataset in each training iteration.
% In addition, we adopt the transductive learning framework in training the DGAT model.
% In terms of the hyperparameter searching, based on our empirical experience, we advice to set the search range for $\epsilon$ between $10^{-7}$ to $10^{-3}$, and search range for $\alpha$ between $0.1$ to $0.9$.
% Moreover, the hyperparameter $\gamma$ in Table \ref{table:hyper} exhibits a correlation between the homophily level:
% the fine-tuned values of $\gamma$ for strong homophily datasets appear to be larger than those strong heterophily datasets.
% This observation suggests that the homophily level of a graph may have an impact on the choice of $\gamma$ for the DGAT model.

\section{Additional Benchmark Results}
\label{app:nine-benchmark}

The benchmark on the 9 small real-world datasets from \cite{pei2020geom},
namely 
\textit{chameleon},
\textit{squirrel},
\textit{actor},
\textit{cornell},
\textit{wisconsin},
\textit{texas},
\textit{cora},
\textit{citeseer} and
\textit{pubmed},
demonstrates that DGAT combines the advantages of GAT and DGN. 
The experiments use 10 fixed splits from \cite{pei2020geom},
which divide nodes into 48\%/32\%/20\%.
We perform a grid search for the hyperparameters of DGN according to \cite{beaini2021directional} for the learning rate in $\{ 10^{-5}, 10^{-4} \}$, the weight decay in $\{ 10^{-6}, 10^{-5}\}$, the dropout rate in $\{0.3, 0.5\}$, the aggregator in \{``mean-dir1-av", ``mean-dir1-dx", ``mean-dir1-av-dir1-dx" \}, the net type in \{``complex", ``simple" \}.
DGAT adopts the same training setting as GAT,
which consists of 2 layers, 8 heads and same hidden size for each datasets.
The hidden size is set to 48 for \textit{chameleon} and \textit{squirrel}, 32 for \textit{actor}, \textit{cornell}, \textit{wisconsin} and \textit{texas}, 16 for \textit{cora} and \textit{citeseer}, and 64 for \textit{pubmed}.
All methods are trained using the Adam optimzer with 1000 steps.
Table \ref{table:hyper} gives the optimal hyperparamters for DGAT on 9 real-world dataset from \cite{pei2020geom}.

Comparing DGAT against GAT and DGN,
the results in Table \ref{table:real-res-vertical} demonstrate that DGAT enhances both graph attention and directional aggregation by incorporating directional information into the computation of graph attention.
We observed that DGAT outperforms GAT on all datasets, 
particularly those with pronounced heterophilic characteristics,
namely,
\textit{chameleon}, 
\textit{squirrel},
\textit{actor},
\textit{cornell},
\textit{wisconsin} and
\textit{texas}.
As shown in \cite{beaini2021directional},
DGN holds advantages over GAT due to its inherent anisotropic nature.
Table \ref{table:real-res-vertical} provides further evidence to support this argument.
In addition,
the results show the superior improvement of DGAT compared to DGN across all homophilic datasets,
namely,
\textit{cora}, \textit{citeseer} and \textit{pubmed}.
This outcome serves as evidence that our proposed model effectively combines the advantages of both mechanisms.
\begin{table}[h!]
 \resizebox{1\hsize}{!}{
  \begin{tabular}{lcccccc}
  \toprule

    Dataset ($\mathbf{H}_{\mathrm{node}}(\mathcal{G})$) &
    % Models\textbackslash{}Hyperparameters &
    learning rate &
    weight decay &
    dropout &
    $\epsilon$  &
    $\gamma$ &
    $\alpha$
    \\

    \midrule
    chameleon (0.25)&
    $5 \times 10^{-3}$ & 
    $5 \times 10^{-4}$ &
    0.4 &
    % $5 \times 10^{-6}$ &
    $1 \times 10^{-7}$ &
    0.0 &
    0.9
    \\
    
    %\midrule
    squirrel (0.22)&
    $5 \times 10^{-2}$ & 
    $5 \times 10^{-5}$ &
    % 0.5 &
    % $5 \times 10^{-6}$ &
    0.2 &
    $1 \times 10^{-7}$ &
    0.2 &
    0.1
    \\

    %\midrule
    actor (0.22)&
    $1 \times 10^{-5}$ & 
    $1 \times 10^{-7}$ &
    % 0.2 &
    % $5 \times 10^{-4}$ &
    0.3 &
    $1 \times 10^{-7}$ &
    0.2 &
    0.2 
    \\
    
    %\midrule
    cornell (0.30) &
    $5 \times 10^{-3}$ & 
    $5 \times 10^{-4}$ &
    0.2 &
    $5 \times 10^{-6}$ &
    0.3 &
    0.8
    \\
    
    %\midrule
    wisconsin (0.16) &
    $5 \times 10^{-3}$ & 
    $5 \times 10^{-6}$ &
    % 0.2 &
    % $5 \times 10^{-6}$ &
    0.5 &
    $1 \times 10^{-6}$ &
    0.3 &
    0.7
    \\
    
    %\midrule
    texas (0.06) &
    $5 \times 10^{-5}$ & 
    $5 \times 10^{-3}$ &
    % 0.4 &
    % $5 \times 10^{-3}$&
    0.4 &
    $1 \times 10^{-5}$&
    0.2 &
    0.8 
    \\
    
    %\midrule
    cora (0.83) &
    $5 \times 10^{-3}$ & 
    $5 \times 10^{-4}$ &
    % 0.5 &
    % $ 1\times 10^{-5}$&
    0.4 &
    $ 5\times 10^{-7}$&
    0.9 &
    0.8 
    \\
    
    %\midrule
    citeseer (0.71)&
    $5 \times 10^{-2}$ & 
    $5 \times 10^{-5}$ &
    % 0.3 &
    0.6 &
    $ 1\times 10^{-5}$&
    0.5 &
    0.2 
    \\
    
    %\midrule
    % {\textbf{PubMed}}
    pubmed (0.79) &
    $1 \times 10^{-2}$ & 
    $5 \times 10^{-2}$ &
    % 0.5 &
    % $ 1\times 10^{-5}$&
    0.6 &
    $ 1\times 10^{-6}$ &
    1 &
    0.3
    \\

  \bottomrule
  \end{tabular}
  }
  \caption{Hyperparameters of DGAT on geom datasets}
  \label{table:hyper}
    
\end{table}
\begin{table}[htbp]
  %\left 
  \resizebox{1\hsize}{!}{
  \begin{tabular}{lccc}
  \toprule 
    Dataset ($\mathbf{H}_{\mathrm{node}}(\mathcal{G})$)
    & $\text{GAT}^{\dag}$
    & DGN  
    & DGAT
    %& $H_{\mathrm{node}}(\mathcal{G})$
    \\
    
    \midrule
    chameleon (0.25)
    & 42.93
    & 50.48 $\pm$ 1.26
    & \textbf{52.47 $\pm$ 1.44}
    %& 0.23
    \\

    squirrel (0.22)
    &30.03
    &\textbf{37.46 $\pm$ 1.28}
    & 34.82 $\pm$ 1.60
    %& 0.22
    \\

    actor (0.22)
    & 28.45
    & 35.47 $\pm$ 1.13
    & \textbf{35.68 $\pm$ 1.20}
    %& 0.22
    \\

    cornell (0.30)
    & 54.32
    & 77.03 $\pm$ 4.72
    & \textbf{85.14 $\pm$ 5.30}
    %& 0.30
    \\

    wisconsin (0.16)
    & 49.41
    & 82.75 $\pm$ 3.01
    & \textbf{84.71 $\pm$ 3.59}
    %& 0.21
    \\

    texas (0.06)
    &58.38
    &78.11 $\pm$ 4.90
    &\textbf{79.46 $\pm$ 3.67}
    %& 0.11
    \\

    cora (0.83)
    &86.37
    &84.69 $\pm$ 1.32
    &\textbf{88.05 $\pm$ 1.09}
    %&0.81
    \\

    citeseer (0.71)
    & 74.32
    & 73.87 $\pm$ 1.32
    & \textbf{76.41 $\pm$ 1.45}
    %& 0.74
    \\

    pubmed (0.79)
    & 87.62
    & 84.29 $\pm$ 0.61
    & \textbf{87.93 $\pm$0.48}
    %& 0.80
    \\
    % just change the previous table into vertical
    
  \bottomrule
  \end{tabular}
  }
  \caption{Benchmark with baselines on nine small real-world datasets.
  The best results are highlighted. 
  Results "$\dag$" are reported from \cite{pei2020geom}. 
  %Results "$\ddag$" are ran by searching the hyperparameters in the similar ranges as Beaini et al.'s experimental setting \cite{beaini2021directional}.
  For fair comparison, we use the same experimental setup as 
  \cite{pei2020geom},
  and report the average test accuracy over the same 10 fixed splits.}
  \label{table:real-res-vertical}
  
\end{table}

\section{Additional Ablation Studies}
\label{app:ablation}
Table \ref{table:ablation_study_vertical} gives the ablation results on the datasets from \cite{pei2020geom}.
The results suggest that the utilization of three proposed mechanisms, namely topology-guided graph rewiring, global directional attention, and parameterized normalized Laplacian matrix, enhances the original GAT. 
Additionally, 
consistent with the earlier observation, graphs exhibiting stronger heterophily characteristics experience greater benefits from the directional message passing approach.
Furthermore, it is worth noting that the use of graph rewiring in DGAT yields remarkable performance enhancements. These improvements contribute to the enhanced performance on the heterophilic datasets, namely \textit{chameleon}, \textit{squirrel}, \textit{actor}, \textit{cornell}, \textit{wisconsin}, and \textit{texas}, resulting in performance gains of $22.22\%$, $15.95\%$, $25.41\%$, $56.74\%$, $71.44\%$ and $36.11\%$ respectively.

% \section{Synthetic Experiment Results}
% \label{app:Synthetic}
% % \begin{landscape}
% % \begin{sidewaysfigure}
% \begin{figure}[h!]
%     \centering
%     \resizebox{1\hsize}{!}{
%     \includegraphics[scale=0.15]{Images/syn-res-new.png}}
%     \caption{Synthetic datasets experiment results. The light and dark green bars represent the accuracies of the DGAT model with $\gamma$ that have the \textit{best} and \textit{worst} results respectively}  
%     \label{fig:syn-res}
% \end{figure}

\end{document}